\documentclass[preprint,12pt,authoryear]{elsarticle}

\usepackage{amssymb,amsmath,amsfonts,bm}
\usepackage{graphicx}
\usepackage{caption}
\usepackage{subcaption}
\usepackage{multirow}
\usepackage{fullpage}
\usepackage{float} 
\usepackage{placeins}
\usepackage[usenames,dvipsnames]{xcolor}

\usepackage[hidelinks]{hyperref}
\hypersetup{colorlinks,bookmarksopen,linkcolor=blue,citecolor=blue,urlcolor=blue}

\newcommand{\revision}[1]{{\color{black}#1}}

\usepackage[normalem]{ulem}

\journal{preprint}

\begin{document}

\begin{frontmatter}



\title{Homogenization with Guaranteed Bounds via Primal-Dual Physically Informed Neural Networks} 


\author[inst1]{Liya Gaynutdinova} 
\author[inst3]{Ond\v rej Roko\v s \corref{cor1}}
\author[inst1]{Martin Do\v sk\' a\v r}
\author[inst1]{Ivana Pultarov\'a}

\affiliation[inst1]{organization={Czech Technical University in Prague},
            addressline={Thákurova 7}, 
            city={Prague},
            postcode={16629}, 
            country={Czech Republic}}

\affiliation[inst3]{organization={Eindhoven University of Technology},
            addressline={5600 MB}, 
            city={Eindhoven},
            postcode={P.O. Box 513},
            country={The Netherlands}}

\cortext[cor1]{Corresponding author, email: liya.gaynutdinova@fsv.cvut.cz}

\begin{abstract}
Physics-informed neural networks (PINNs) have shown promise in solving partial differential equations (PDEs) relevant to multiscale modeling, but they often fail when applied to materials with discontinuous coefficients, such as media with piecewise constant properties. This paper introduces a dual formulation for the PINN framework to improve the reliability of the homogenization of periodic thermo-conductive composites, for both strong and variational (weak) formulations. The dual approach facilitates the derivation of guaranteed upper and lower error bounds, enabling more robust detection of PINN failure. We compare standard PINNs applied to smoothed material approximations with variational PINNs (VPINNs) using both spectral and neural network-based test functions. Our results indicate that while strong-form PINNs may outperform VPINNs in controlled settings, they are sensitive to material discontinuities and may fail without clear diagnostics. In contrast, VPINNs accommodate piecewise constant material parameters directly but require careful selection of test functions to avoid instability. Dual formulation serves as a reliable indicator of convergence quality, and its integration into PINN frameworks enhances their applicability to homogenization problems in micromechanics.
\end{abstract}


\end{frontmatter}

\section{Introduction}

Accurately modelling the effective behaviour of materials with complex microstructures is crucial in various scientific and engineering domains. Simulation of 
components made of such materials, while resolving all microstructural details, makes multi-scale modeling prohibitively computationally demanding for practical everyday use. Homogenization aims to replace a heterogeneous material with a homogeneous one that exhibits an equivalent macroscopic behaviour \citep{Engquist2008}. For most computational homogenization problems, physics-informed neural networks (PINNs) \citep{RAISSI2019686} currently cannot match the computational efficiency and reliability of classical Finite Element Method (FEM) and Fast Fourier Transform (FFT) approaches \citep{GEERS20102175, Davoli}, despite promising theoretical advantages \citep{Grossmann}. However, PINNs demonstrate clear advantages for mesh-free computation of complex microstructures, multi-query parametric studies, and scenarios involving irregular geometries where traditional meshing becomes problematic \citep{RODRIGUES201873}.

Recent advances in PINN architectures have begun addressing fundamental limitations in multiscale problems \citep{Park2022, Leung2022, Soyarslan2024}. These developments, combined with the inherent advantages of mesh-free computation and rapid evaluation for trained networks, position PINNs as valuable tools for specific applications while classical methods remain optimal for computational homogenization workflows in production.

A critical gap remains in establishing rigorous error bounds and robust formulations for elliptic partial differential equations (PDEs) within the PINN framework. While existing PINN approaches primarily rely on the strong form of the PDE residual, this formulation can be sensitive to solution regularity and may not provide reliable error estimates \citep{krishnapriyan2021, deryck2023, mishra2023}. Variational PINNs (VPINNs) address some of these limitations by incorporating weak formulations that utilize test functions and integrate the PDE residual over the domain, thereby imposing weaker regularity requirements and potentially enhancing solution stability. However, VPINNs remain sensitive to the choice of test functions and still lack robustness in error estimation \citep{E2018, kharazmi2019}.

To address these challenges in PINN-based homogenization, we propose a scheme that enables the calculation of guaranteed upper and lower bounds for effective parameters. Using \revision{two microstructure examples}, we demonstrate how classic PINNs can fail when applied to piecewise constant materials. Although sophisticated techniques exist for handling discontinuous materials, including GFEM \citep{Legrain}, X-FEM \citep{Belytschko_2009}, the immersed boundary method \citep{LeVeque}, domain decomposition PINNs \citep{AmeyaDJagtap2020}, and various adaptive PINN approaches \citep{Liu2023, ZHAO2025118184}, robust application of advanced PINNs in homogenization settings remains under development.

Our approach employs both primal and dual formulations of the elliptic PDE governing the homogenization problem, rather than relying solely on the primal formulation. The dual formulation provides complementary information that enables more accurate and reliable bounds on effective material properties, as demonstrated for classical homogenization methods by \cite{Gaynutdinova2022, Gaynutdinova2023}. For the network architecture, we adopt the frameworks of \cite{JIANG2023115972} and \cite{Soyarslan2024}, which incorporate periodic boundary conditions directly without loss penalty terms. We extend their architecture to solve the dual formulation problem while satisfying the divergence-free constraint \citep{richterpowell2022, farkane2023}, then use the primal and dual solutions to derive error bounds. Recognizing that the strong form can still yield poor performance in classic PINNs, we also implement the primal-dual approach for VPINNs and compare results across different test function choices \citep{Berrone2022, ROJAS2024116904}. Our results reveal that strong-form PINNs outperform both VPINNs and FEM benchmarks only when materials exhibit sufficiently gradual phase transitions, while VPINNs show sensitivity to test function selection. In all cases, however, the guaranteed bounds to the effective parameters can be utilized to assess the quality of the computed results.

The rest of this work is structured as follows. In Section \ref{sec:theory}, we recall the zeroth-order homogenization theory for both primal and dual formulations in the steady-state linear 2D heat equation. In Section \ref{sec:NN}, we introduce the PINN architecture and construct the strong training loss for the dual formulation, as well as the variational loss for both primal and dual formulations. \revision{In Subsection \ref{sec:PINN}, we compare the performance of the classic PINNs for different material functions and address a common mode of PINN failure, whereas in Subsection \ref{sec:VPINN} we compare the performance of VPINNs for different types and cardinalities of test bases; in both cases, for a single inclusion. Subsection~\ref{sec:multiple_inc} then extends this discussion to a microstructure with multiple inclusions.} Section \ref{sec:discussion} summarizes our findings and discusses the pros and cons of each method, and Section \ref{sec:conclusion} provides the outlook for future applications.

\section{Primal and Dual Formulation of Homogenization}\label{sec:theory}

In this part, we employ computational homogenization, which determines the effective properties of heterogeneous materials by solving boundary value problems on their representative periodic cells. First, we introduce the governing equation for the homogenization of linear steady-state heat transport in 2D by recalling the strong form of the equation, from which the typical PINN loss is constructed \citep{Park2022, Leung2022, JIANG2023115972, WU2023112521, Soyarslan2024}, and introduce its dual form. Then we present the variational forms for both the primal and dual formulations, which are used in most classical numerical methods, and will serve as the basis for the VPINN approach. Finally, we show how to calculate the estimates of the homogenized conductivity matrix, as well as its guaranteed upper and lower bounds.

\subsection{Strong form}\label{sec:strong_form}

Let us consider a periodic unit cell $X \subset \mathbb{R}^2$ representing a periodic heterogeneous solid, where its microstructural thermal conductivity second-order tensor \revision{can be represented in the matrix notation as a $2 \times 2$ symmetric matrix $\bm{A}$. Its spatial variability within the periodic cell is then expressed as a function of the spatial coordinate~$\bm{x}$ as} $\bm{A}(\bm{x}):X \to \mathbb{R}_{\rm{ sym}}^{2\times 2}$, which is uniformly positive definite, with each component essentially bounded in $X$. According to the zeroth-order homogenization theory \citep{MICHEL1999109, jikov2012homogenization}, the temperature field $u(\bm{x})$ within the unit cell takes the form
\begin{equation}\label{eq:U}
    u(\bm{x}) = \bm{\xi}^{\rm{T}}\bm{x} + \widetilde{u}(\bm{x}),\;\bm{x} \in X,
\end{equation}
where $\bm{\xi}^{\rm{T}}\bm{x}$ represents the averaged contribution for the prescribed macroscopic temperature gradient $\bm{\xi}$ and $\widetilde{u}(\bm{x})$ is the micro-fluctuation part caused by the material heterogeneity. The temperature fluctuation $\widetilde{u}(\bm{x})$ is a $X$-periodic function in the $x_1$ -- $x_2$ plane, $\bm{x}=[x_1, x_2]$. 

Under steady-state conditions \revision{with no heat sources}, the heat flux $\bm{j}(\bm{x})$ is related to the temperature gradient $\bm{e}(\bm{x}) = \nabla u(\bm{x}) = [\frac{\partial u}{\partial x_1}, \frac{\partial u}{\partial x_2}]^{\rm T}$ by Fourier’s law as:
\begin{equation}\label{eq:q}
    \bm j (\bm{x})= - \bm{A}(\bm{x}) \bm{e}(\bm{x})=- \bm{A}(\bm{x}) \nabla u(\bm{x}),
\end{equation}
\revision{and the divergence of the heat flux must be zero, i.e.,
\begin{equation}\label{eq:j_zero}
    \nabla \cdot \bm j (\bm{x})= 0.
\end{equation}
}

The strong form of the governing differential equation of the steady-state heat conduction then requires one to find $\widetilde {u}\in C^2_{\rm per}(X)$ such that:
\begin{equation}\label{eq:primal_strong}
    \nabla \cdot \left[\bm{A}(\bm{x}) ({\bm{\xi}}+ \nabla \widetilde{u}(\bm{x})) \right] = 0, \; \forall \bm{x}\in X,
\end{equation}
where $\nabla \cdot$ denotes the divergence operator, \revision{i.e., }$\nabla \cdot \bm{f}=\frac{\partial f_1}{\partial x_1}+\frac{\partial f_2}{\partial x_2}$\revision{ and $C^2_{\rm per}(X)$ denotes the space of twice-differentiable periodic continuous function on $X$}.

The dual form of Eq.~\eqref{eq:primal_strong} is defined for the thermal resistivity $\bm{A}^{-1}(\bm{x})$ \citep{VONDREJC2015}. For the heat transfer problem, we can derive it analogously from the property of the temperature gradient field $\bm{e}$ \citep{briane2008duality, milton2022theory}, following the Helmholtz decomposition theorem \citep{jikov2012homogenization}, i.e.:
\begin{equation}\label{eq:e}
    \nabla \times \bm{e}(\bm{x}) = \bm{0},  \; \forall \bm{x}\in X.
\end{equation}
The vector field $\bm{e}(\bm{x})$ then relates back to the heat flux through the resistivity tensor:
\begin{equation}\label{eq:e_to_q}
    \bm e (\bm{x})= - \bm{A}^{-1}(\bm{x}) \bm{j}(\bm{x}).
\end{equation}
We can again decompose the heat flux $\bm{j}(\bm{x})$ into a constant part $\bm{\zeta}$ and a fluctuating part $\widetilde{\bm{v}}(\bm{x})$, resulting in the dual form of \eqref{eq:primal_strong} as
\begin{equation}\label{eq:dual_strong_v}
        \nabla \times \left[\bm{A}^{-1}(\bm{x}) (\bm{\zeta} + \widetilde{\bm{v}}(\bm{x})) \right] = \bm{0},\; \forall \bm{x}\in X. 
\end{equation}
Importantly, this new vector field $\widetilde{\bm{v}}(\bm{x})$ has to satisfy the divergence-free condition of the admissible heat flux fields, i.e.:
\begin{equation}\label{eq:div_free}
        \nabla \cdot \widetilde{\bm{v}}(\bm{x}) = 0.
\end{equation}
In the two-dimensional setting, every divergence-free vector function $\widetilde{\bm{v}}(\bm{x})$ can be derived from some periodic scalar function $\widetilde{w}(\bm{x})$ (the so-called stream function) as  
\begin{equation}\label{eq:curl}
    \widetilde{\bm{v}}(\bm{x}) = \bm{Q} \nabla  \widetilde{w}(\bm{x}),
\end{equation}
where $\bm{Q}$ is a $90^{\circ}$ rotation matrix \revision{in} the $x_1$--$x_2$ plane \citep{Girault1979, milton2022theory} \revision{ of the form
\begin{equation}\label{eq:Q}
    \bm{Q} = \begin{bmatrix}
                0 & -1 \\
                1 & 0
             \end{bmatrix}.
\end{equation}
The divergence of vector $ \widetilde{\bm{v}}(\bm{x})$ can then be written as 
\begin{equation}\label{eq:v}
     \nabla \cdot \widetilde{\bm{v}}(\bm{x}) = \left[\frac{\partial}{\partial x_1}, \frac{\partial}{\partial x_2}\right] \cdot 
     \begin{bmatrix}
        -\frac{\partial \widetilde{w}}{\partial x_2} \\
        \frac{\partial \widetilde{w}}{\partial x_1} 
     \end{bmatrix}
     = -\frac{\partial^2 \widetilde{w}}{\partial x_2 \partial x_1} + \frac{\partial^2 \widetilde{w}}{\partial x_1 \partial x_2},
\end{equation}
which is always equal to zero, thanks to the symmetry of partial derivatives.}
Then, the problem \eqref{eq:dual_strong_v} can be written as to find $\widetilde{w}(\bm{x}) \in C^2_{\rm per}(X)$ such that
\begin{equation}\label{eq:dual_strong}
    \nabla \times \left[\bm{A}^{-1}(\bm{x}) (\bm{\zeta} + \bm{Q} \nabla \widetilde{w}(\bm{x})) \right] = \bm{0},\; \forall \bm{x}\in X, 
\end{equation}
\revision{with} the divergence-free constraint \eqref{eq:div_free} is automatically fulfilled. Note that \revision{in order to apply} the curl operator $\nabla \times $ \revision{to} a 2D vector function $\bm{f}$, \revision{the function has to be} expanded to a 3D vector function $[f_1,f_2,f_3 = 0]$. The condition $\nabla \times \bm{f} = \bm{0}$ can then be written as
\begin{equation}
    \left[ \frac{\partial f_3}{\partial x_2} - \frac{\partial f_2}{\partial x_3}, 
    \frac{\partial f_1}{\partial x_3} - \frac{\partial f_3}{\partial x_1}, 
    \frac{\partial f_2}{\partial x_1} - \frac{\partial f_1}{\partial x_2} \right] = \bm{0}.
\end{equation}
Since $f_3=0$ and $\bm{f}$ is a function in the $x_1$--$x_2$ plane, the first two components are always zero, and the equation \eqref{eq:dual_strong} simplifies to contain only the last component, thus reducing to a scalar problem\revision{~similar to~Eq.~(\ref{eq:primal_strong})}. We keep the original notation for brevity, but from now on, when we write $\nabla \times \bm{f}$, we actually mean its scalar form $\frac{\partial f_2}{\partial x_1} - \frac{\partial f_1}{\partial x_2}$.

\subsection{Weak form}\label{sec:weak_form}
Despite promising theoretical advantages, strong-form PINNs currently cannot match the computational efficiency and reliability of classical computational approaches for most PDEs \citep{Grossmann}. As the analytical solution of problems \eqref{eq:primal_strong} and \eqref{eq:dual_strong} is usually unavailable, most frameworks, such as the finite element method (FEM) \citep{yvonnet2019computational} and Fourier-Galerkin method \citep{moulinec1994fast, schneider2021review}, rely on their respective variational (weak) forms. This allows for the relaxation of the solution regularity requirement from $C^2_{\rm per}(X)$ 
\revision{to space $H^1_{\rm per}(X) = \{\bm{u}\in L^2(X,\mathbb{R}); \bm{u}_{\rm per}\in H^1_{\rm loc}(\mathbb{R}^2,\mathbb{R}) \}$, 
where $\bm{u}_{\rm per}$ is the $X$-periodic extension of $\bm{u}$ and $H^1_{\rm loc}(\mathbb{R}^2,\mathbb{R})$ are functions with squared gradients integrable on every compact subset of $\mathbb{R}^2$.}  Instead of solving \eqref{eq:primal_strong}, we want to find $\widetilde{u}\in H^1_{\rm per}(X)$ such that
\begin{equation}\label{eq:primal_weak}
      \int_X (\nabla \phi(\bm{x}))^{\rm{T}} \bm{A}(\bm{x}) (\bm{\xi} + \nabla \widetilde u( \bm{x})) {\rm d}\bm{x} = 0, \;\; \text{for all } \phi \in H^1_{\rm per}(X),
\end{equation}
where $\phi(\bm{x})$ is a test function. \revision{The boundary term $ \int_{\partial X}(n \cdot \nabla \phi) \widetilde u( \bm{x}) {\rm d}s=0$, which appears alongside~(\ref{eq:primal_weak}) after the integration by parts,  vanishes automatically, since it has opposite values on the opposite boundaries
of $X$ due to periodicity.}
Similarly, the variational form of \eqref{eq:dual_strong} is to find $\widetilde{w}\in H^1_{\rm per}(X)$ such that
\begin{equation}\label{eq:dual_weak}
     \int_X (\bm{Q}\nabla\psi(\bm{x}))^{\rm{T}}\; \bm{A}^{-1}(\bm{x}) (\bm{\xi} + \bm{Q} \nabla \widetilde w( \bm{x})) {\rm d}\bm{x} = 0, \;\; \text{for all } \psi \in H^1_{\rm per}(X),
\end{equation}
with a test function $\psi(\bm{x})$\revision{$\in H^1_{\rm per}(X)$}. 

\subsection{Upper and lower bound to homogenized parameters}
The exact solution of \eqref{eq:primal_weak} is also the \revision{minimizer} of the related homogenization problem~\citep{jikov2012homogenization}, which reads to find
a homogenized coefficient matrix $\bm{A}^*\in {\mathbb R}^{2\times 2}$ such that
\begin{equation}\label{eq:min1}
\langle \bm{A}^*\bm{\xi}, \bm{\xi} \rangle_{{\mathbb R}^2} = \inf_{\widetilde{u} \in H^1_{\rm per}(X)} \frac{1}{\vert X\vert}
\int_X \langle  \bm{A}(\bm{x}) (\bm{\xi}+\nabla \widetilde{u}(\bm{x})),\bm{\xi}+\nabla \widetilde{u}(\bm{x}) \rangle_{{\mathbb R}^2}\, {\rm d}\bm{x}, \; \forall \bm{\xi} \in {\mathbb R}^2,
\end{equation}
where $\vert X\vert$ denotes the area of the domain $X$, and $\langle \bm{u}, \bm{v}\rangle_{\mathbb{R}^2}$ denotes the inner product of
vectors $\bm{u}$ and $\bm{v}$ of the Euclidean space $\mathbb{R}^2$.

Conversely, the inverse $\bm{B}^*$ of the homogenized coefficient $\bm{A}^*$ can be calculated using the dual formulation \citep{haslinger1995optimum, VONDREJC2015, Gaynutdinova2022}:
\begin{equation}\label{eq:min2}
\langle \bm{B}^*\bm{\zeta}, \bm{\zeta} \rangle_{{\mathbb R}^2} = \inf_{\widetilde{w} \in H^1_{\rm per}(X)}\frac{1}{\vert X\vert}
\int_X \langle  \bm{A}^{-1}(\bm{x}) (\bm{\zeta}+\bm{Q}\nabla\widetilde{w} (\bm{x})),\bm{\zeta}+\bm{Q}\nabla \widetilde{w} (\bm{x}) \rangle_{{\mathbb R}^2} {\rm d}\bm{x}, \; \forall \bm{\zeta} \in {\mathbb R}^2.
\end{equation}

Given the variational structure of \eqref{eq:min1} and \eqref{eq:min2}, \revision{for which the true minimizer exists in the infinite-dimensional space $H^1_{\rm per}(X)$, substituting any minimizer over subspaces of $H^1_{\rm per}(X)$ (i.e., approximate solutions $\widetilde{u}_h$ and $\widetilde{w}_h$)} into the integrals in \eqref{eq:min1} and \eqref{eq:min2}, respectively, for  given $\bm{\xi}$ and $\bm{\zeta}$ yields larger values \revision{of the effective coefficients} than for the true minimizers, and thus \revision{it yields }the upper bounds $\bm{A}^*_h$ and $\bm{B}^*_h$ to $\bm{A}^*$ and $\bm{B}^*$ in the sense that
\begin{equation}\label{eq:order}
    (\bm{B}^*_h)^{-1} \preceq (\bm{B}^*)^{-1} = \bm{A}^* \preceq \bm{A}^*_h,
\end{equation}
where the partial ordering $\bm{K} \preceq \bm{L}$ means that $\bm{v}^{\rm T} \bm{K} \bm{v} \leq \bm{v}^{\rm T} \bm{L} \bm{v}$ for all $\bm{v} \in \mathbb{R}^2$.

In practice, we solve \eqref{eq:primal_strong} or \eqref{eq:primal_weak} (as well as \eqref{eq:dual_strong} and \eqref{eq:dual_weak}) for given $\bm{\xi}$ and $\bm{\zeta} \in \{[1,0]^{\rm T}, [0,1]^{\rm T}\}$ to compute the diagonal elements of $\bm{A}^*_h$ and $\bm{B}^*_h$.

\section{Physically Informed Neural Networks for Homogenization}\label{sec:NN}

\begin{figure}
    \centering
    \includegraphics[width=1\linewidth]{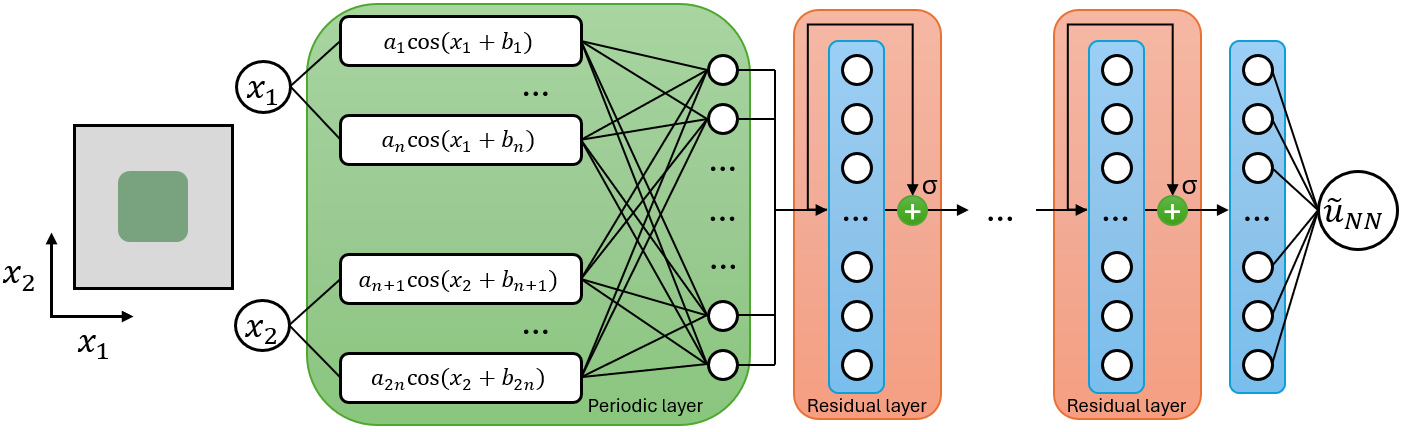}
    \caption{Scheme of the primal PINN architecture introduced by \cite{JIANG2023115972}. The network inputs the microscopic coordinates $[x_1, x_2]$ and outputs the fluctuating temperature $\widetilde{u}_{\rm NN}$. Identical architecture is used for the dual formulation, where the output is the stream function $\widetilde{w}_{\rm NN}$.}
    \label{fig:PINN_scheme}
\end{figure}

To show how existing PINN homogenization approaches can be improved by introducing the dual form training loss, we adopt the PINN architecture introduced by \cite{JIANG2023115972} (Fig.~\ref{fig:PINN_scheme}), which was demonstrated to produce a good approximation of the solution of \eqref{eq:primal_strong} for a wide range of different material functions with smooth coefficients, see Fig.~\ref{fig:PINN_scheme}. This network is characterized by a unique periodic layer that employs learnable cosine functions. This design choice ensures that the network's output is inherently periodic, thereby eliminating the need to incorporate a boundary condition term into the training loss.

\FloatBarrier

The network takes as input the microscopic coordinates $x_1$ and $x_2$ and produces output in one of two forms: it can either deliver the fluctuating temperature $\widetilde{u}_{\rm NN}$ in the primal form or provide the stream function $\widetilde{w}_{\rm NN}$ in the dual form. \revision{Thus, by training two separate networks with the same architecture, we can obtain a primal and a dual solution to the homogenization problem. The architecture of each network includes the periodic layer alongside one or more sequential residual layers, which utilize smooth activation functions like the hyperbolic tangent.}

\subsection{Strong form}

The PDE residual from \eqref{eq:primal_strong} or \eqref{eq:dual_strong} is used as the PINN training loss, i.e.:

\begin{equation}\label{eq:primal_loss}
    \mathcal{L}_{\text{p,s}} := \frac{1}{|X|} \int_X \left( \nabla \cdot \left[\bm{A}(\bm{x}) ({\bm{\xi}}+ \nabla \widetilde{u}_{\rm NN}(\bm{x})) \right] \right)^2 \text{d}\bm{x},
\end{equation}
for the primal strong-form PINN, and
\begin{equation}\label{eq:dual_loss}
    \mathcal{L}_{\text{d,s}} := \frac{1}{|X|} \int_X \left( \nabla \times \left[\bm{A}^{-1}(\bm{x}) (\bm{\zeta} + \bm{Q}\nabla\widetilde{w}_{\rm NN}(\bm{x})) \right] \right)^2 \text{d}\bm{x},
\end{equation}
for the dual strong-form PINN.

\subsection{Weak form}

In this paper, we also aim to address certain limitations of PINNs for the homogenization of piecewise constant materials that are often encountered in the laminate or matrix-inclusion forms. Because the strong formulation is used in training the networks, the material function $\bm{A}(\bm{x})$ is also assumed to have continuous first-order gradients. While the material smoothness condition is not strictly necessary for the technical implementation of PINNs, ignoring it often leads to poor results \citep{HENKES2022114790}, so an artificial smoothing of the material function is usually employed. Moreover, in our particular homogenization setting, naively implementing piecewise-constant material coefficients leads to failure, as the network converges to a near-constant solution because the term $\nabla \cdot (\bm{A}(\bm{x}) {\bm{\xi}})$ is zero in all possible collocation points. Introducing material smoothing, however, alters the underlying problem and introduces a material approximation error into the estimates of the homogenized parameter. We later show that even for smooth functions, an insufficiently gradual transition between the material phases can cause poor PINN solutions, because $\nabla \cdot (\bm{A}(\bm{x}) {\bm{\xi}})$ becomes very large at the interface between the materials and very small elsewhere. 

A potential remedy lies in replacing the strong form of the governing PDE with its weak form, which allows for the relaxation of the smoothness constraint in both the material function and the solution. \revision{While there are many sophisticated methods for addressing material discontinuities \citep{Liu2023,lei2025}}, we aim to test the limits of the standard approach by letting the VPINN have an identical architecture to the PINN described above. It is important to note that the weak formulation also broadens the options for activation functions; however, to isolate the impact of the formulation itself, we will continue to use the hyperbolic tangent activation function. 

We implement the loss function for Robust VPINNs introduced by \cite{ROJAS2024116904}. For the primal neural network solution $\widetilde{u}_{\rm NN}$ and a basis of test functions $\phi_n(\bm{x}) \in H^1_{\rm per}(X)$, $n=1,\dots,N_t$ we compute the vector of residuals $\bm{r}_{\rm p}(\widetilde{u}_{\rm NN})=[r_{{\rm p}}^1, \dots, r_{{\rm p}}^{N_t}]^{\rm T}$:
\begin{equation}\label{eq:weak_residual_primal}
    r_{{\rm p}}^n(\widetilde{u}_{\rm NN}) = \int_X (\nabla \phi_n(\bm{x}))^{\rm{T}} \; \bm{A}(\bm{x}) (\bm{\xi} + \nabla \widetilde u_{\rm NN}(\bm{x})) {\rm d}\bm{x}, \; n=1,\dots,N_t.
\end{equation}
As shown in \citep{ROJAS2024116904}, minimizing the sum of squared residuals with respect to normalized individual test functions is equivalent to minimizing the functional
\begin{equation}\label{eq:weak_loss}
    \mathcal{L}_{\text{p,w}} := \bm{r}_{\rm p}^{\rm{T}}(\widetilde{u}_{\rm NN}) \bm{G}^{-1} \bm{r}_{\rm p}(\widetilde{u}_{\rm NN}),
\end{equation}
where $\bm{G}$ is a symmetric positive definite Gram matrix, $G_{nm}=\int_X (\nabla \phi_n)^{\rm T} \nabla\phi_m \, {\rm d}\bm{x}$. 

The dual weak loss functional is defined analogously with the test functions $\psi_n(\bm{x}) \in H^1_{\rm per}(X)$, $n=1,\dots,N_t$, where the weak dual residual is 
\begin{equation}\label{eq:weak_residual_dual}
r^n_{{\rm d}}(\widetilde{w}_{\rm NN}) =\int_X (\bm{Q} \nabla \psi_n(\bm{x}))^{\rm{T}} \bm{A}^{-1}(\bm{x}) (\bm{\zeta} + \bm{Q} \nabla\widetilde w_{\rm NN}(\bm{x})) {\rm d}\bm{x}, \; n=1,\dots,N_t.
\end{equation}

\section{Numerical Experiments}
This section demonstrates how training PINNs in the dual formulation of the problem can complement the solution in the primal form, and compares the effectiveness of VPINNs against classic PINNs \revision{for unit cells with a singular (Section~\ref{sec:single_inclusion}) and multiple (Section~\ref{sec:multiple_inc}) inclusions}.

\subsection{Singular Inclusion}
\label{sec:single_inclusion}
\begin{figure}[h]
    \centering
    \includegraphics[width=\linewidth]{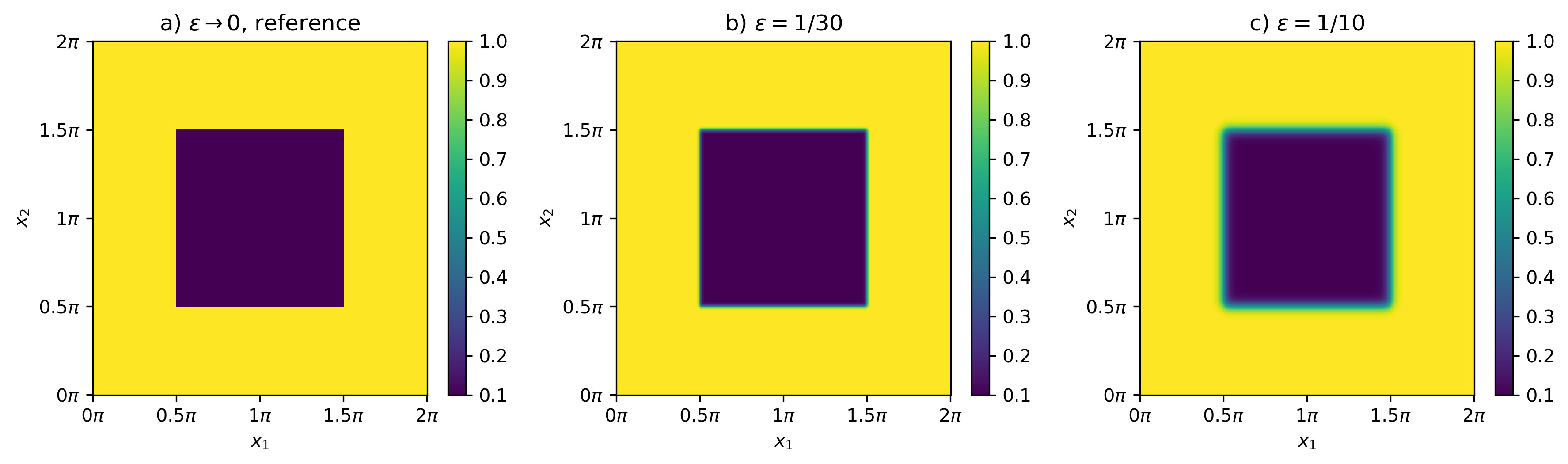}
    \caption{Smooth approximations $\bm{A}_{s,\varepsilon}(\bm{x})$ of the material distribution depending on the parameter $\varepsilon$.}
    \label{fig:smooth_material} 
\end{figure}

Consider a square domain $X$ of the size $2\pi \times 2\pi$ with a centrally placed square inclusion of the size $\pi \times \pi$ with isotropic material phases, shown in Fig.~\ref{fig:smooth_material}. For this case, the effective conductivity can be computed exactly \citep{Obnosov} as
\begin{equation}\label{eq:exact}
  \bm{A}^*=\gamma^{\mathrm{eff}} \bm{I} = 
  	\gamma_{\mathrm{mat}} \sqrt{\frac{ \gamma_{\mathrm{mat}}+3 \gamma_{\mathrm{inc}}}{3\gamma_{\mathrm{mat}}+ \gamma_{\mathrm{inc}}}} \;
  	 \bm{I},
\end{equation}
where $\gamma_{\mathrm{mat}}$ and $\gamma_{\mathrm{inc}}$ are thermal conductivities of the matrix and inclusion, respectively. With $\gamma_{\mathrm{mat}}=1$ and $\gamma_{\mathrm{inc}}=0.1$, formula \eqref{eq:exact} yields $\gamma^{\mathrm{eff}} \approx 0.6476$.

\begin{figure}[h]
    \centering
    \includegraphics[width=0.8\linewidth]{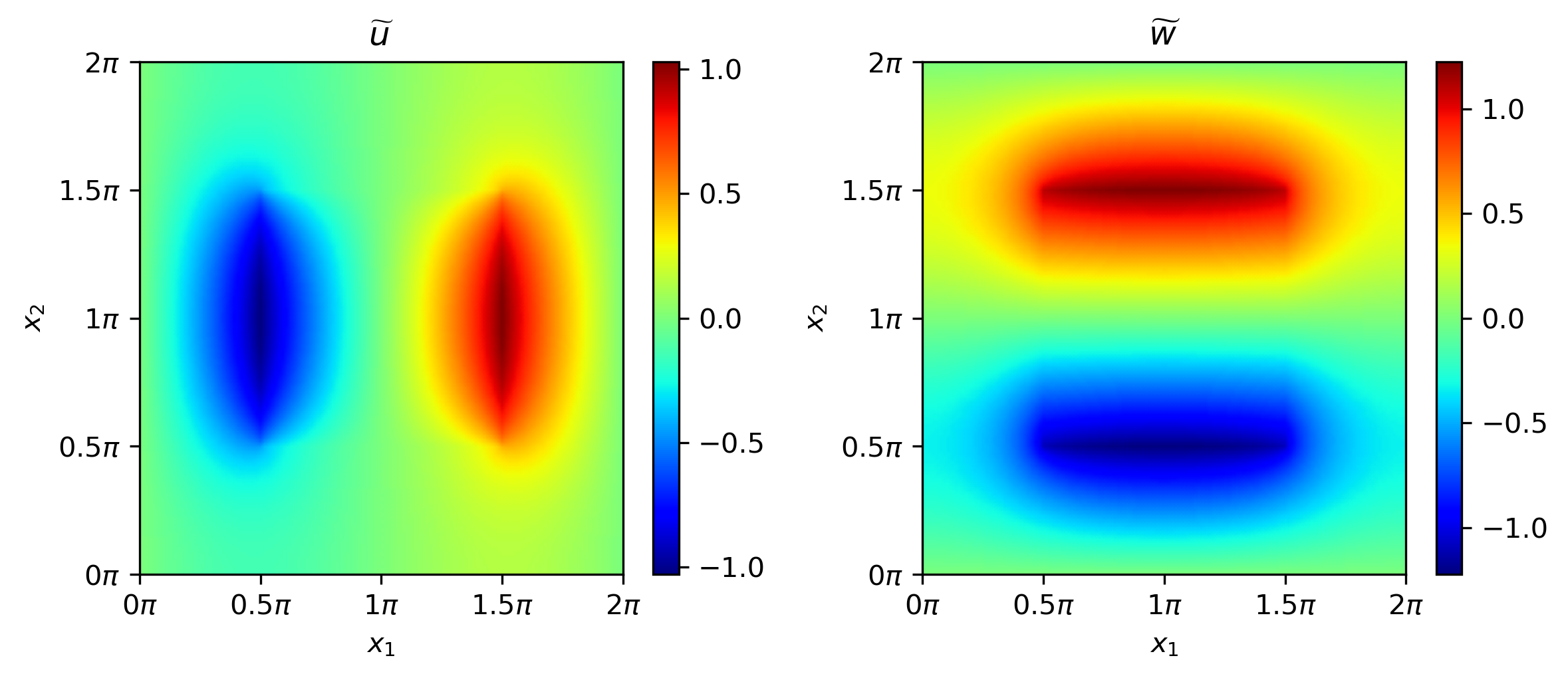}
    \caption{Benchmark primal and dual solutions for $\bm{\xi}=[1,0]^{\rm{T}}$, obtained with FEM, using discretization into $128 \times 128$ unique DoFs.}
    \label{fig:fem} 
\end{figure}

Together with the analytical exact homogenized conductivity $\gamma^{\mathrm{eff}}$, we use a FEM solution for the benchmark on $\widetilde{u}$ and $\widetilde{w}$ (cf.~Fig.~\ref{fig:fem}). The primal and dual FEM solutions provide guaranteed upper and lower bounds to $\gamma^{\mathrm{eff}}$ \citep{Gaynutdinova2022}. \revision{For the sake of simplicity,} we use a regular grid with triangular elements and continuous piecewise linear functions. Since the material distribution is symmetric, we only have to compute the solution for one macroscopic gradient ($\bm{\xi},\bm{\zeta}=[1,0]^{\rm{T}}$), and we will be comparing the estimates of the first diagonal element of the homogenized heat conductivity matrix only. For the solution on a regular mesh containing $128 \times 128$ unique degrees of freedom (DoFs) ($129 \times 129$ grid with periodic boundary nodes), these bounds are very close to the exact analytical effective conductivity (within $\pm 0.04 \%$ of relative error). 

In all experiments, we consider four neural network architectures of varying parameter sizes, as described in Tab.~\ref{tab:NN_config} for both primal and dual formulations. 
\revision{To reduce the impact of the choice of the collocation points, we train all the networks using the same grid as the highest resolution Finite Element Method (FEM) solution, which has 16,384 collocation points. These points are then utilized for numerical integration with the trapezoidal rule to calculate the training losses in Eqs.~\eqref{eq:primal_loss}--\eqref{eq:weak_residual_primal} and \eqref{eq:weak_residual_dual}. They are also used for effective parameter estimation in Eqs.~\eqref{eq:min1} and \eqref{eq:min2}.}

\begin{table}[h]
    \centering
    \begin{tabular}{|c|c|c|c|}
        \hline
        \footnotesize{\textnumero~neurons, periodic layer} & \footnotesize{\textnumero~neurons, per \revision{residual} layer} & \footnotesize{\textnumero~\revision{residual} layers} & \footnotesize{\textnumero~parameters}\\
        \hline
        4 & 4 & 1 & 65\\
        \hline
        10 & 10 & 2 & 391\\
        \hline
        20 & 20 & 3 & $1\,801$\\
        \hline
        50 & 50 & 5 & $15\,601$\\
        \hline
    \end{tabular}
    \caption{Configurations of the considered neural networks for both primal/dual and strong/weak formulations, \revision{single inclusion}.}
    \label{tab:NN_config}
\end{table}

\subsubsection{Strong form}\label{sec:PINN}

As discussed earlier, problems with piecewise constant material distribution cannot be solved with the classic PINNs based on the strong formulation. 
As a workaround, we adopt a smooth ($C^{\infty}_{\rm per}(X)$) approximation of the piecewise constant material function, similarly to \cite{JIANG2023115972}. In this work, the approximation takes the form
\begin{equation}
    \bm{A}_{s,\varepsilon}(\bm{x}) = [1 - 0.9 \cdot p_{\varepsilon}(x_1) \cdot p_{\varepsilon}(x_2)] \; \bm{I},
\end{equation}
where
\begin{equation}
    p_{\varepsilon}(x_i) = \frac{1}{2} \left( 1+\tanh \left[\frac{1}{\varepsilon} \, \sin \left(x_i-\frac{\pi}{2} \right) \right] \right),
\end{equation}
and where parameter $\varepsilon > 0$ controls the steepness of the material transition; smaller $\varepsilon$ leads to a narrower interpolation area between the two materials, see examples in Fig.~\ref{fig:smooth_material}. 

\begin{figure}[h]
    \centering
    \includegraphics[width=1\linewidth]{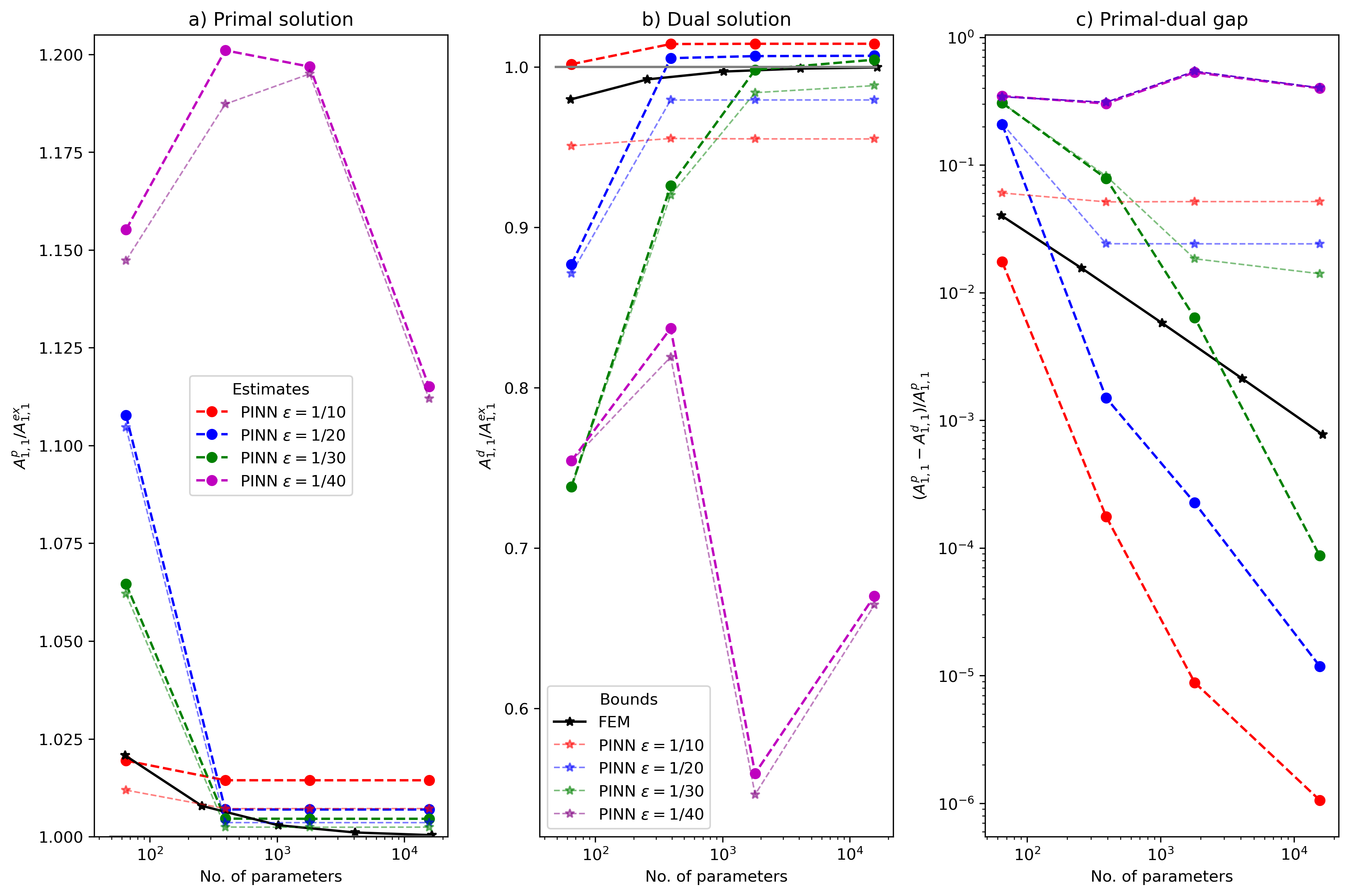}
    \caption{Comparison of the primal and dual estimates (-$\bullet$-) and guaranteed bounds (-$\star$-) of the PINN solutions depending on the number of network parameters and the transition parameter $\varepsilon$ of the material approximation.}
    \label{fig:pinn-estimates}
\end{figure}

To establish the link between the smoothness of the material and the solution quality, we trained all network architectures of Tab.~\ref{tab:NN_config} on material distributions with $\varepsilon \in \{1/10, 1/20, 1/30, 1/40\}$ for $40\,000$ epochs \revision{with the Adam optimizer and the starting learning rate $10^{-5}$}. The networks output the primal and dual solutions, which can be used to compute estimates of the homogenized conductivity, as well as the guaranteed upper and lower bounds. For the estimates, we approximate the integrals in \eqref{eq:min1} and \eqref{eq:min2} by calculating the terms $\nabla \widetilde{u}(\bm{x})$ and $\bm{Q} \nabla \widetilde w( \bm{x})$ in the collocation points, and then using trapezoidal rule. Since these gradients must be calculated during training, \revision{the gap between the primal and dual estimates provides a fast additional convergence} metric to follow during the training process. The \revision{final estimates} are summarized in Fig.~\ref{fig:pinn-estimates} as dots connected by bold dashed lines. Here we can see that these estimates can be quite precise, i.e., have a very low relative difference $(A_{1,1}^p-A_{1,1}^d)/A_{1,1}^p$ between them (i.e., primal-dual gap), especially for the wider material transition ($\varepsilon=1/10, 1/20$). However, the order of Eq.~\eqref{eq:order} is not guaranteed for these quick estimates, due to the imprecision of the integration \citep{haslinger1995optimum}. Furthermore, the estimates only hold for the smooth material distribution $\bm{A}_{s,\varepsilon}(\bm{x})$, which introduces its own approximation error, and the networks converge to an effective conductivity around 0.5\% higher than the analytical solution. 

\revision{In principle, the computation of the guaranteed bounds requires two conditions to be satisfied: (i) the integrals in Eqs.~\eqref{eq:min1} and ~\eqref{eq:min2} are calculated precisely; and (ii) the approximate solutions $\widetilde{u}(\bm{x})$ and $\widetilde{w}(\bm{x})$ are from the subspaces of $H^1_{\rm per}(X)$. To tackle the challenge of precisely integrating the PINN solution (since it can be difficult to derive a closed form), we project it onto finite-element functions, which also form a subspace of $H^1_{\rm per}(X)$. The simplest approach is to set the projected solution in the nodes of some finite element mesh to be equal to the PINN values at these points and interpolate linearly between them. As a result, the terms $\nabla \widetilde{u}(\bm{x})$ and $\bm{Q} \nabla \widetilde w( \bm{x})$ remain constant within each triangular subdomain. If the material distribution is piecewise constant and the mesh is designed to align with the material phases, both the primal and dual flux will also be constant in each triangular subdomain. This makes the exact integration of the function trivial. Here, we use the same regular mesh that was used for the FEM benchmark computations, which also aligns with the training collocation points.}

The resulting bounds to the conductivity parameter are denoted by stars connected by thin lines in  Fig.~\ref{fig:pinn-estimates}. Because the chosen material approximation increases the effective conductivity, the bounds obtained from the solution with $\epsilon \in \{1/10,1/20,1/30\}$ are closer to each other and to the exact value. On the other hand, the lower bounds are actually below the exact value, i.e., they are guaranteed, unlike the ``quick'' estimates.

Generally, the primal PINN estimates converge slightly faster for an increasing number of parameters than their dual counterpart, unlike the FEM primal and dual solutions, which converge at the same rate. For smaller $\varepsilon$, larger networks are required to achieve the same primal-dual gap. The primal-dual gap decreases the fastest for the material with the most gradual transition ($\varepsilon=1/10$), at the price of wider guaranteed bounds, when the actual discontinuous material is considered. The PINN with the lowest \revision{number of} DoFs can yield an estimate on par with the FEM benchmark if the material transition is sufficiently gradual. The best result with respect to the analytical effective conductivity is achieved for the largest network of 15,601 parameters and $\varepsilon=1/30$. 
On the other hand, even large networks produce poor solutions for the considered $\varepsilon=1/40$ and smaller.  

\begin{figure}[h]
    \centering
    \includegraphics[width=1\linewidth]{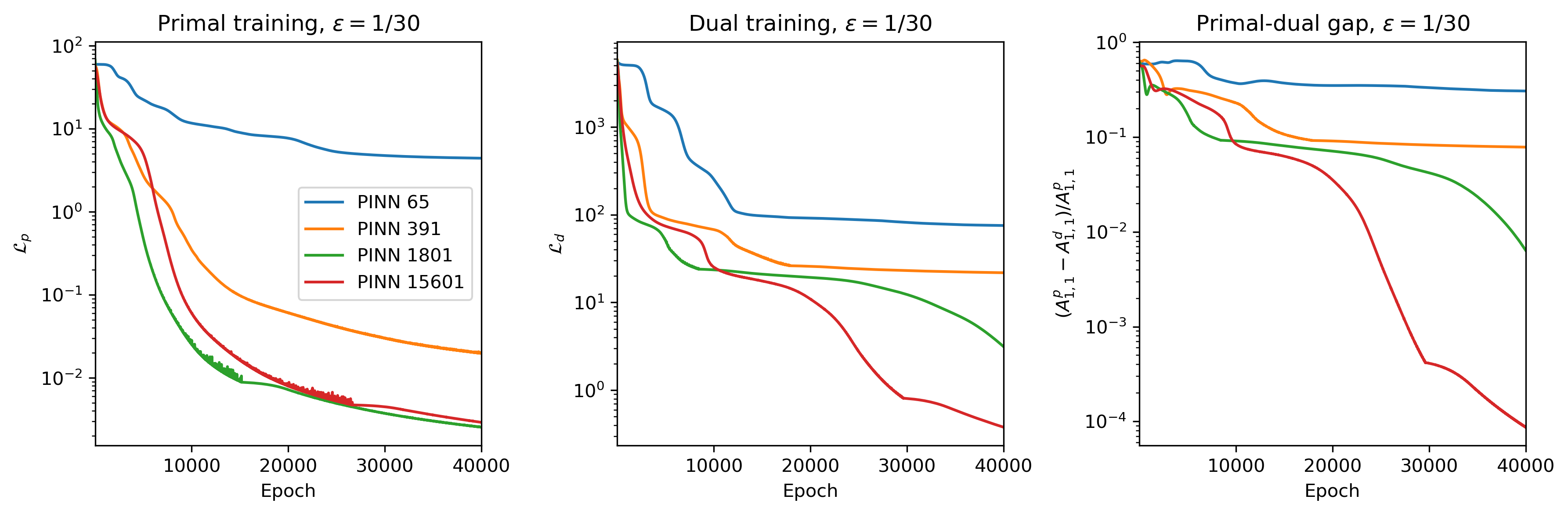}
    \caption{Training loss of the primal and dual solutions and the gap between the estimates, for the PINNs with the transition parameter  $\varepsilon=1/30$.}
    \label{fig:training-pinn30}
\end{figure}

\begin{figure}[h]
    \centering
    \includegraphics[width=1\linewidth]{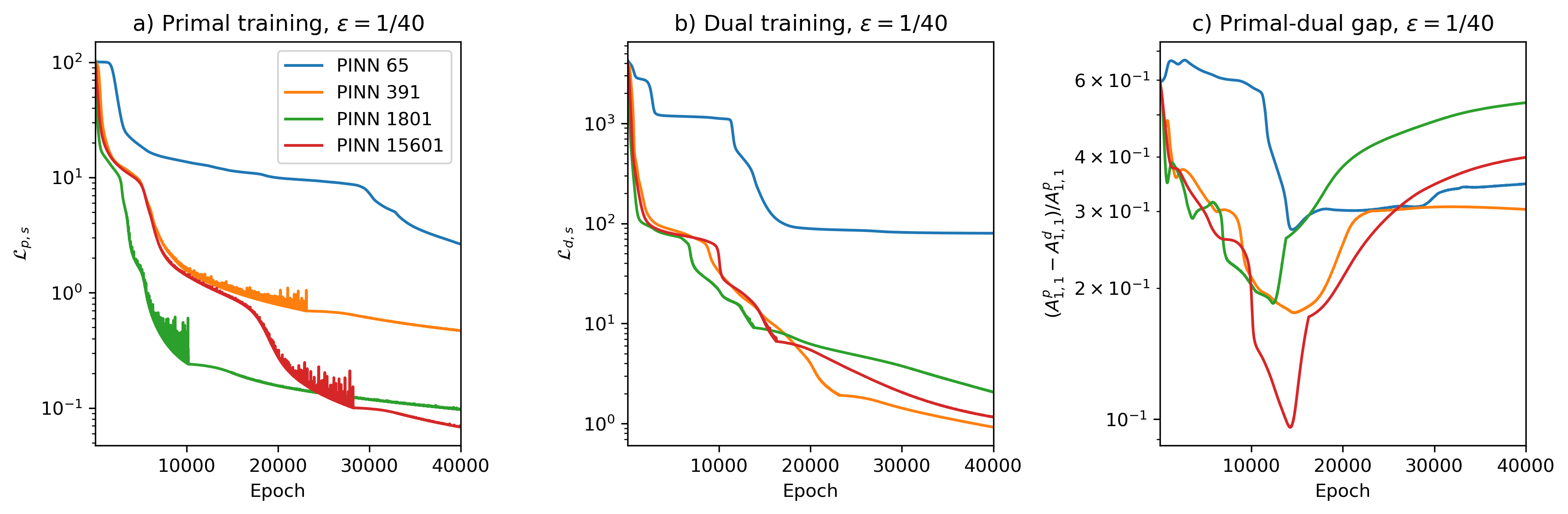}
    \caption{Training loss of the primal and dual solutions and the gap between the estimates, for the PINNs with the transition parameter $\varepsilon=1/40$.}
    \label{fig:training-pinn40}
\end{figure}

\begin{figure}[h]
    \centering
    \includegraphics[width=1\linewidth]{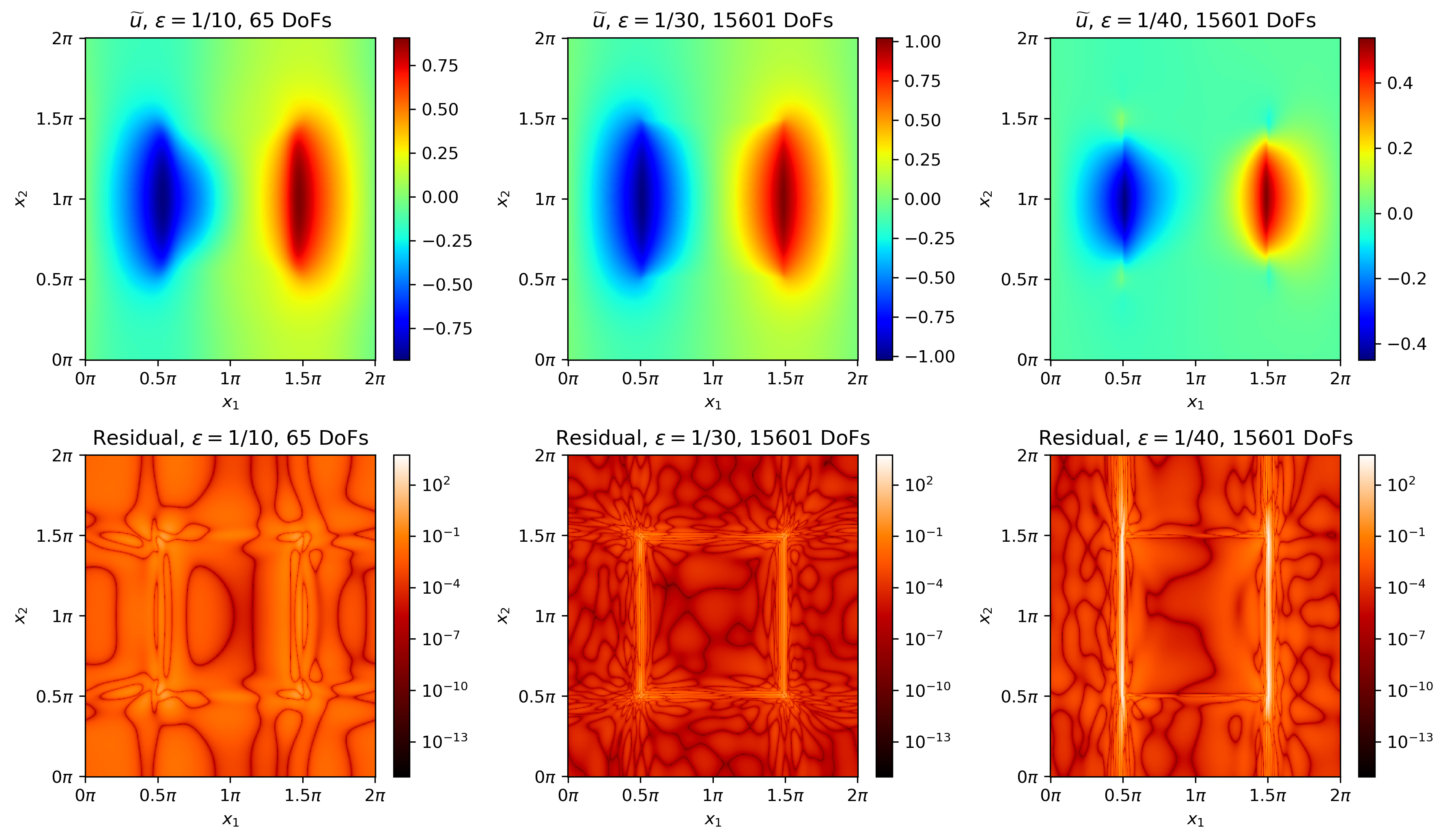}
    \caption{\revision{Primal solutions (top) and the residuals (bottom) of select PINNs depending on the transition parameter $\varepsilon$.}}
    \label{fig:pinn_primal}
\end{figure}

\begin{figure}[h]
    \centering
    \includegraphics[width=1\linewidth]{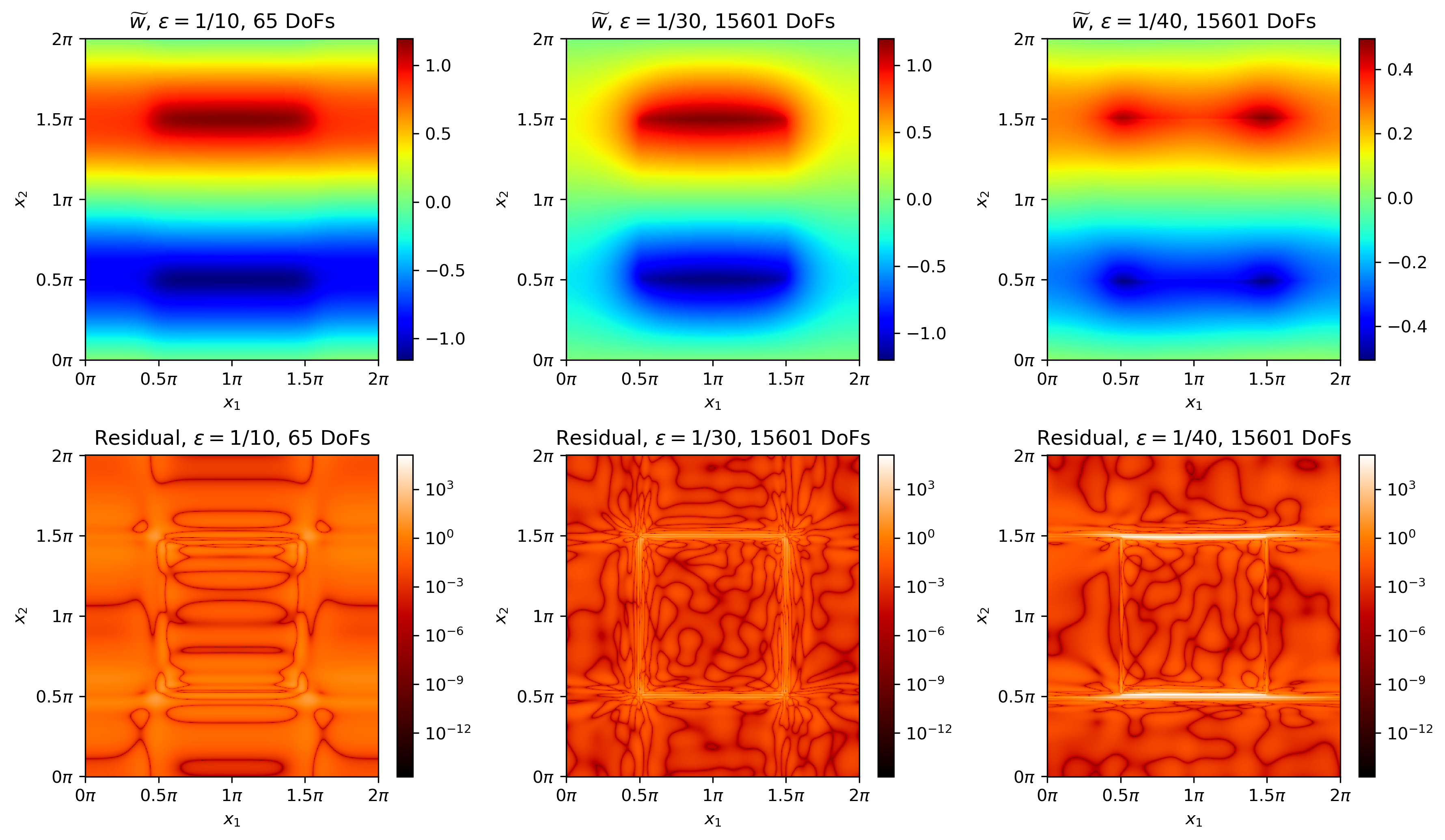}
    \caption{\revision{Dual solutions (top) and the residuals (bottom) of select PINNs depending on the transition parameter $\varepsilon$.}}
    \label{fig:pinn_dual}
\end{figure}

The results reveal that the greatest danger in utilizing only primal PINNs for solving piecewise constant materials (or even their smooth approximations) is the inability to identify \revision{when the achieved solution is poor}. When solely looking at the training process (Figs.~\ref{fig:training-pinn30}--\ref{fig:training-pinn40}) and the resulting residual fields (Figs.~\ref{fig:pinn_primal}--\ref{fig:pinn_dual}), the results might not indicate poor convergence. In particular, compare training of the PINNs for $\varepsilon=1/30$ (Fig.~\ref{fig:training-pinn30}) and $\varepsilon=1/40$ (Fig.~\ref{fig:training-pinn40}), where the training curves look similar, but the achieved \revision{parameter estimates differ} substantially. In the absence of the reference solution, it can also be hard to identify the point of failure, as the residuals are even lower in the area outside the material transition for the $\varepsilon=1/40$ solution than for the $\varepsilon=1/30$ solution, see Figs.~\ref{fig:pinn_primal} and \ref{fig:pinn_dual}. Only the difference between the primal and dual estimates reveals the poor quality of the PINN solutions.

\FloatBarrier

\subsubsection{Weak form}
\label{sec:VPINN}

The inability of the strong-form PINNs to handle discontinuous material distribution can be remedied by considering the weak form of the underlying PDE in the Variational PINNs (VPINNs). For a direct comparison of their performance to the PINNs in numerical homogenization, we consider the same network architectures as in Section~\ref{sec:PINN}, and we use the collocation points on the same $128 \times 128$ grid for evaluating the integrals in Eqs.~\eqref{eq:primal_weak} and \eqref{eq:dual_weak}. As we expect the weak formulation to handle the piecewise constant material, we only train for the original, piecewise constant coefficient matrix $\bm{A}(\bm{x})$. However, in VPINNs, the choice of appropriate test functions becomes crucial \citep{Berrone2022}. Here, we investigate two types of test function bases: a spectral basis (paragraph \textit{VSPINN} below) and a neural network basis (paragraph \textit{VNPINN} below).

\paragraph{\textbf{VPINN with spectral test functions (VSPINN)}}\label{sec:VSPINN}

\begin{figure}[h]
    \centering
    \includegraphics[width=1\linewidth]{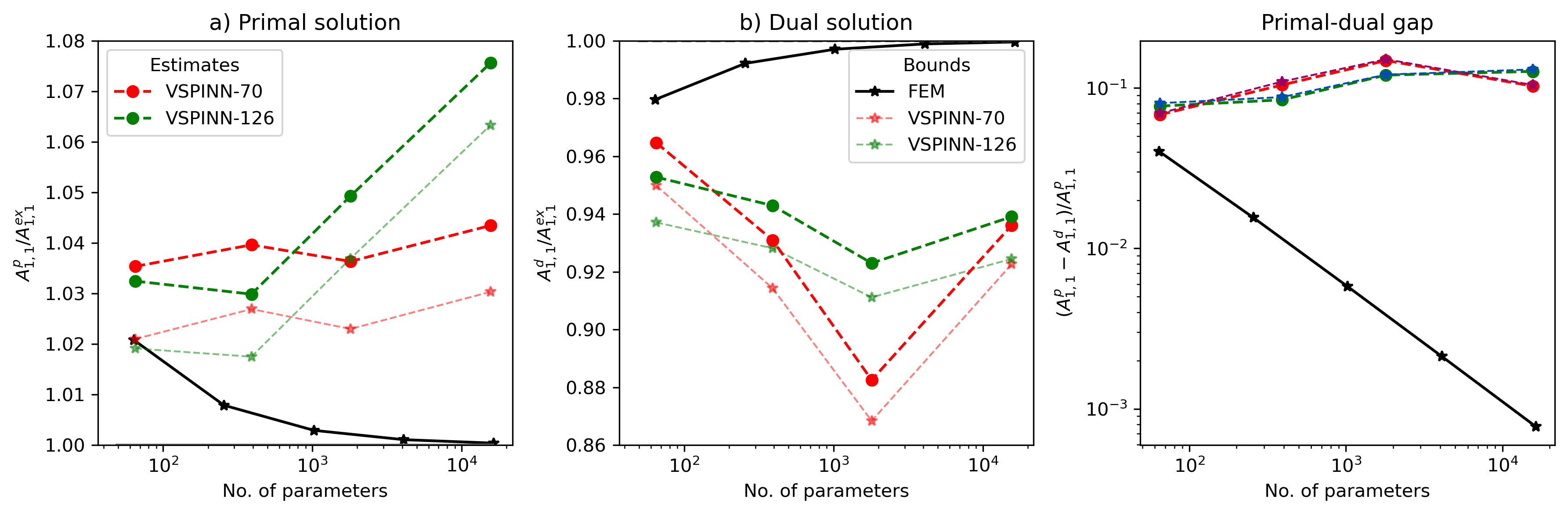}
    \caption{Comparison of the primal and dual estimates (-$\bullet$-) and guaranteed bounds (-$\star$-) of the VSPINN solutions depending on the number of network parameters and the number of spectral test functions.}
    \label{fig:vspinn-estimates}
\end{figure}

First, we consider a typical basis for the periodic square domain $Y=(0,2\pi)\times (0,2\pi)$ derived from the Fourier transform in the form:
\begin{align*}\label{eq:spectral_basis}
    \phi_{s,m,n} &:= \sin(mx_1+nx_2),\\
    \phi_{c,m,n} &:= \cos(mx_1+nx_2),
\end{align*}
where $m = 0, 1, \dots M$, $n = 0, 1, \dots N$, with $(m, n) \ne (0,0)$. The Gram matrix 
$\bm{G}$ of Eq.~\eqref{eq:weak_loss} is then diagonal and thus easily invertible. Each diagonal element of the inverse Gram matrix corresponding to the sine or cosine test function with parameters $m$ and $n$ can be calculated as
\begin{equation}\label{eq:G_spectral}
    \text{diag} \, (G^{-1}_{s,m,n}) = \text{diag} \, (G^{-1}_{c,m,n})=\frac{1}{m^2+n^2}.
\end{equation}
The primal and dual solutions were trained with a basis of size $N_t=70$ spectral test functions ($M=N=5$) and $N_t=126$  functions ($M=N=7)$ with the architecture setups from Tab.~\ref{tab:NN_config}. The resulting estimates and bounds are shown in Fig.~\ref{fig:vspinn-estimates}. 

\begin{figure}[h]
    \centering
    \includegraphics[width=1\linewidth]{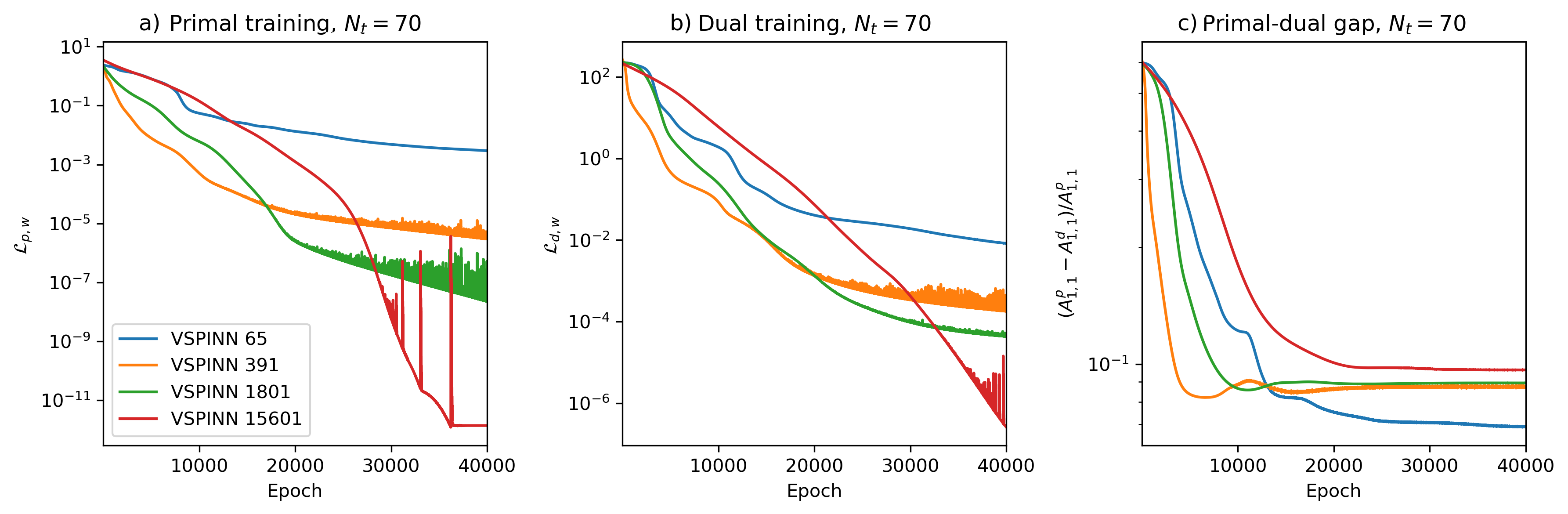}
    \caption{Training loss of the primal and dual solutions, and the gap between the estimates, VPINN with 70 spectral test functions (VSPINN).}
    \label{fig:training70-vspinn}
\end{figure}

\begin{figure}[h]
    \centering
    \includegraphics[width=1\linewidth]{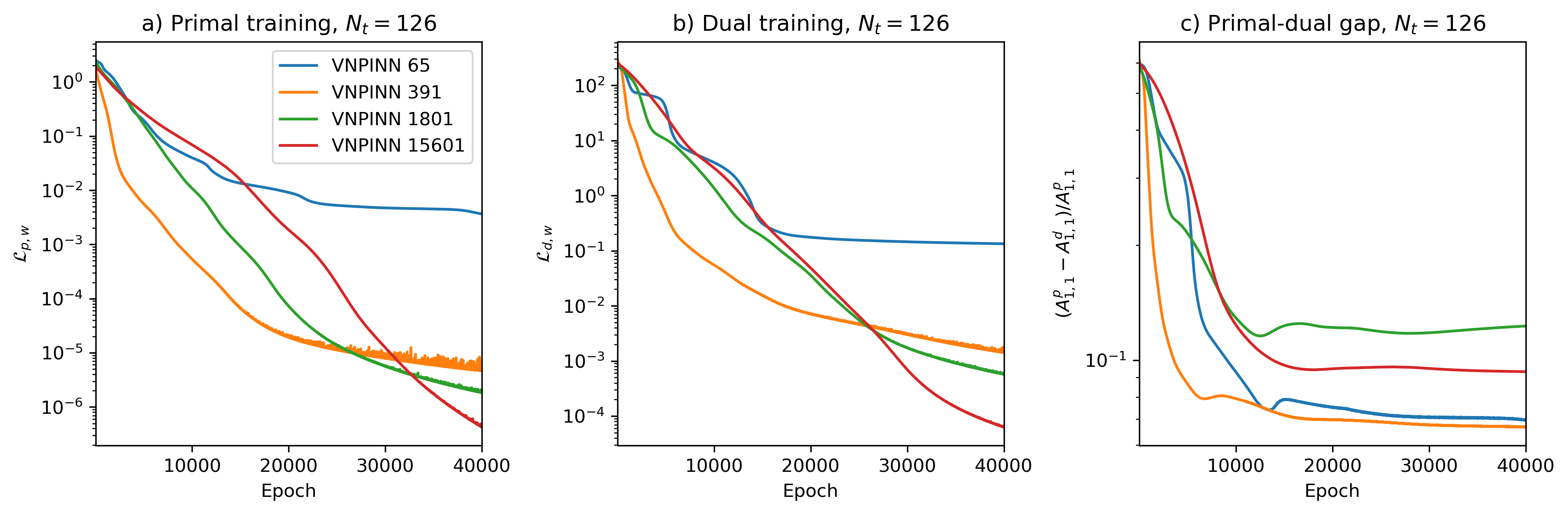}
    \caption{Training loss of the primal and dual solutions, and the gap between the estimates, VPINN with 126 spectral test functions (VSPINN).}
    \label{fig:training126-vspinn}
\end{figure}

Here we can see that the variational PINNs with the spectral test functions (VSPINNs) can handle the discontinuity in the material and yield more precise estimates than PINNs with the small transition phase $\varepsilon = 1/40$. However, larger networks are not more precise than smaller networks, and the performance of the large ($1\,801$ and $15\,601$ parameters) networks with 126 test functions is even worse than the performance of the same network architectures trained with 70 test functions, especially in the primal formulation. In both cases, the training loss of the larger networks was orders of magnitude lower than the training loss of the smaller networks, see Figs.~\ref{fig:training70-vspinn} and \ref{fig:training126-vspinn}. This, however, does not translate into better estimates. 

\begin{figure}[h]
    \centering
    \includegraphics[width=1\linewidth]{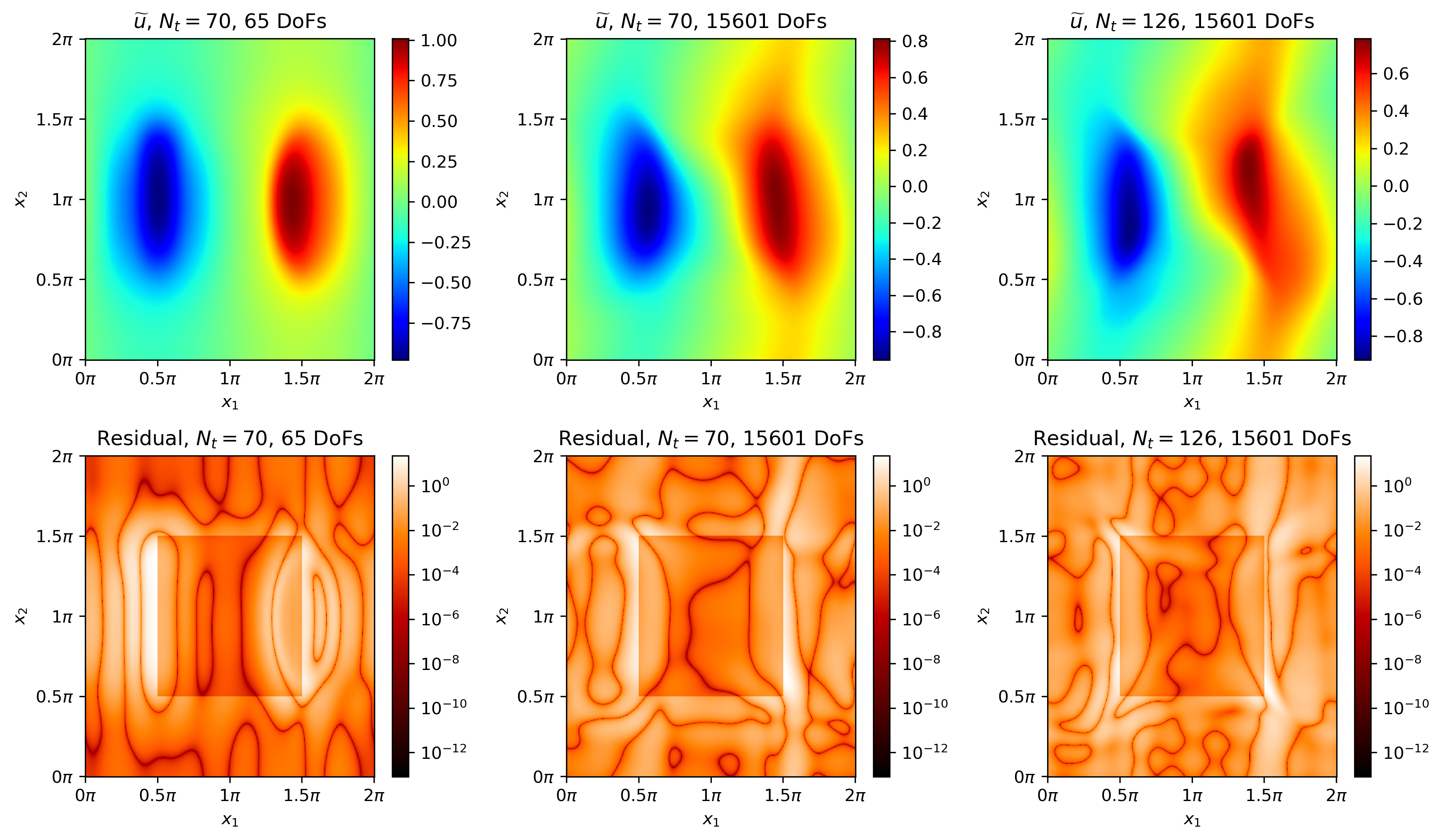}
    \caption{Primal solutions and the residuals of select VSPINNs depending on the number of the spectral test functions $N_t$.}
    \label{fig:vspinn-primal}
\end{figure}

\begin{figure}[h]
    \centering
    \includegraphics[width=1\linewidth]{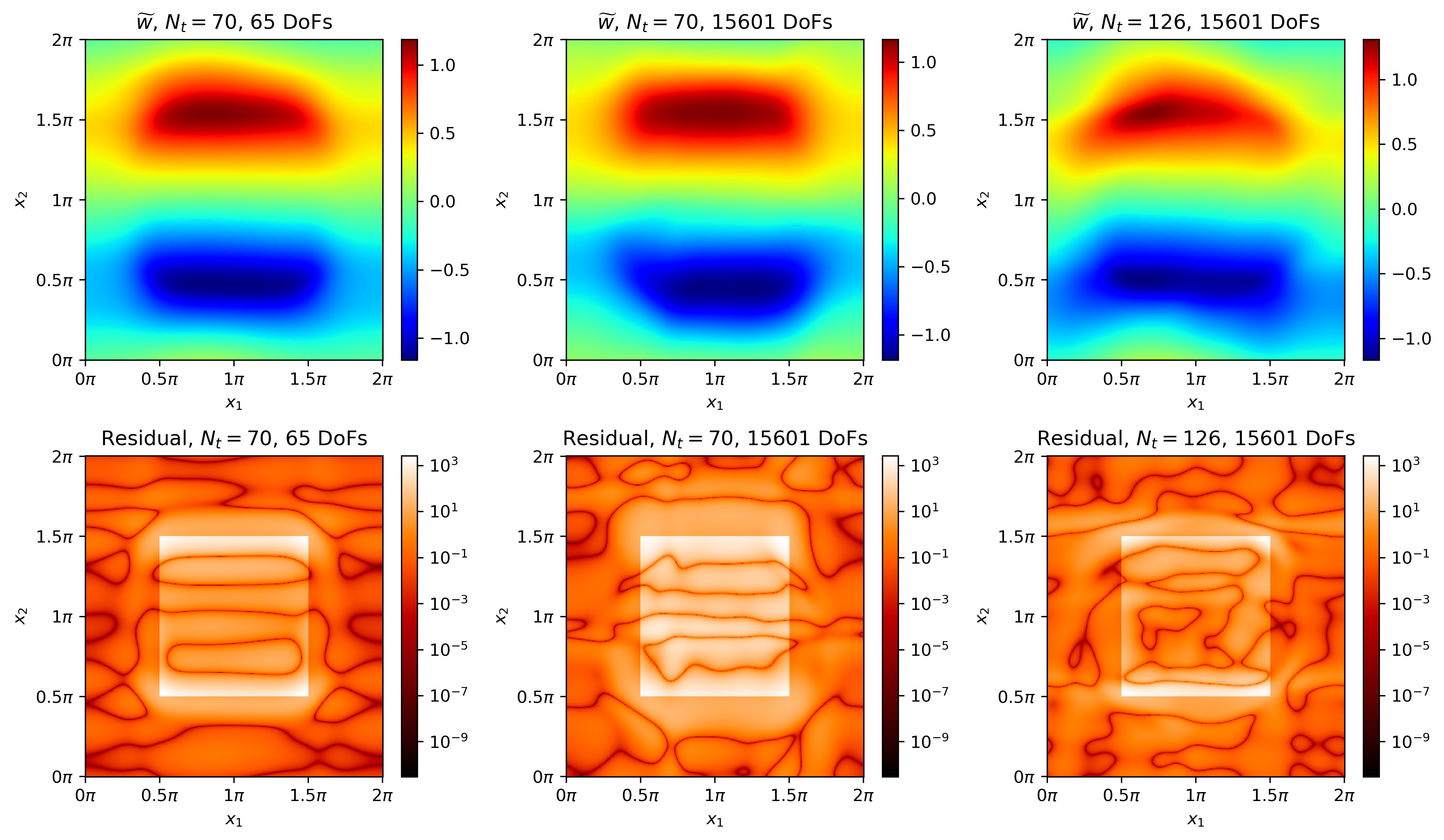}
    \caption{Dual solutions and the residuals of select VSPINNs depending on the number of the spectral test functions $N_t$.}
    \label{fig:vspinn-dual}
\end{figure}

Because the VPINNs use the same architecture as PINNs, the pointwise strong-form residual can be calculated for comparison, shown in Figs.~\ref{fig:vspinn-primal} and \ref{fig:vspinn-dual}. The results suggest that the larger networks are prone to overfitting to the inprecision in the integration of the scalar products with the test functions in \eqref{eq:primal_weak} and \eqref{eq:dual_weak}, especially considering the distortion in the solution with $15\,601$ parameters. This problem is worse for the basis of 126 test functions, as it contains functions with higher frequencies, for which the same integration scheme yields progressively lower precision. Integration problems might be remedied with different integration techniques, such as Filon's quadrature \citep{Chase}, although these are ultimately out of the scope of this paper.

While the weak formulation improves the performance of the smaller networks, allowing them to handle the piece-wise constant material without artificial smoothing, VSPINNs cannot compete with the FEM benchmark. Classic PINNs provide better results for larger networks, even considering the material approximation error. In terms of computational efficiency, there is no clear difference between VPINNs and classic PINNs, as VPINNs eliminate the need to compute the second-order gradients, but at the same time, they require more memory to process the integration with the test functions.

\FloatBarrier

\paragraph{\textbf{VPINN with neural network test functions (VNPINN)}}\label{sec:VNPINN}

\begin{figure}[h]
    \centering
    \begin{subfigure}[b]{\textwidth}
        \centering
        \includegraphics[width=1\linewidth]{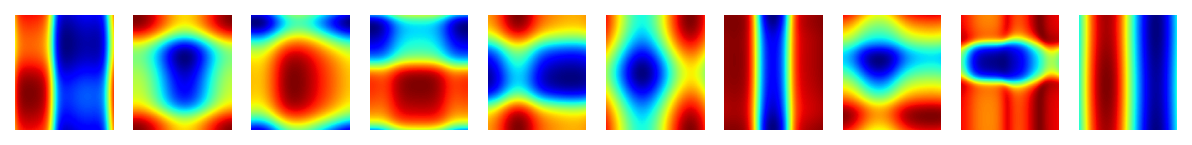}
        \caption{65 parameters}
    \end{subfigure}%

    \begin{subfigure}[b]{\textwidth}
        \centering
        \includegraphics[width=1\linewidth]{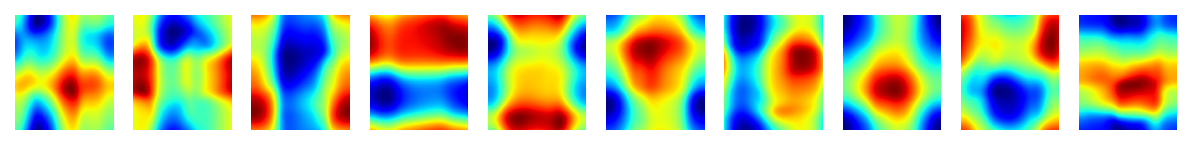}
        \caption{15,601 parameters}
    \end{subfigure}
    \caption{Example of 10 randomly initialized neural network test functions based on the PINN architecture.}
    \label{fig:nn-basis}
\end{figure}

Due to the limitations of spectral basis functions, we explore other options for periodic functions, specifically neural networks. For each considered architecture, several instances are initialised with random weights, see examples in Fig.~\ref{fig:nn-basis}.

The benefits of this approach are twofold: (i) the functions are naturally from the same function space as the solution; (ii) more test functions can be added without worsening the precision of integration.  The main downside is that the linear independence of these test functions is not guaranteed, mainly because of the weight-space symmetry of the neural networks \citep{brea2019}. The \revision{generally non-diagonal} Gram matrix and its inverse in \eqref{eq:weak_loss} have to be computed numerically, and we commonly observe that for the number of test functions approaching the number of neural network parameters, this matrix is no longer positive definite, typically with a few very small negative eigenvalues. Another option would be to construct an orthogonal basis by training the networks, but this is very time-consuming in practice. Instead, when the full Gram matrix was ill-conditioned, particularly for the network with 65 DoFs and $N_t=100$ or 200, \revision{we set all its non-diagonal values to zero.} 

\begin{figure}[h]
    \centering
    \includegraphics[width=1\linewidth]{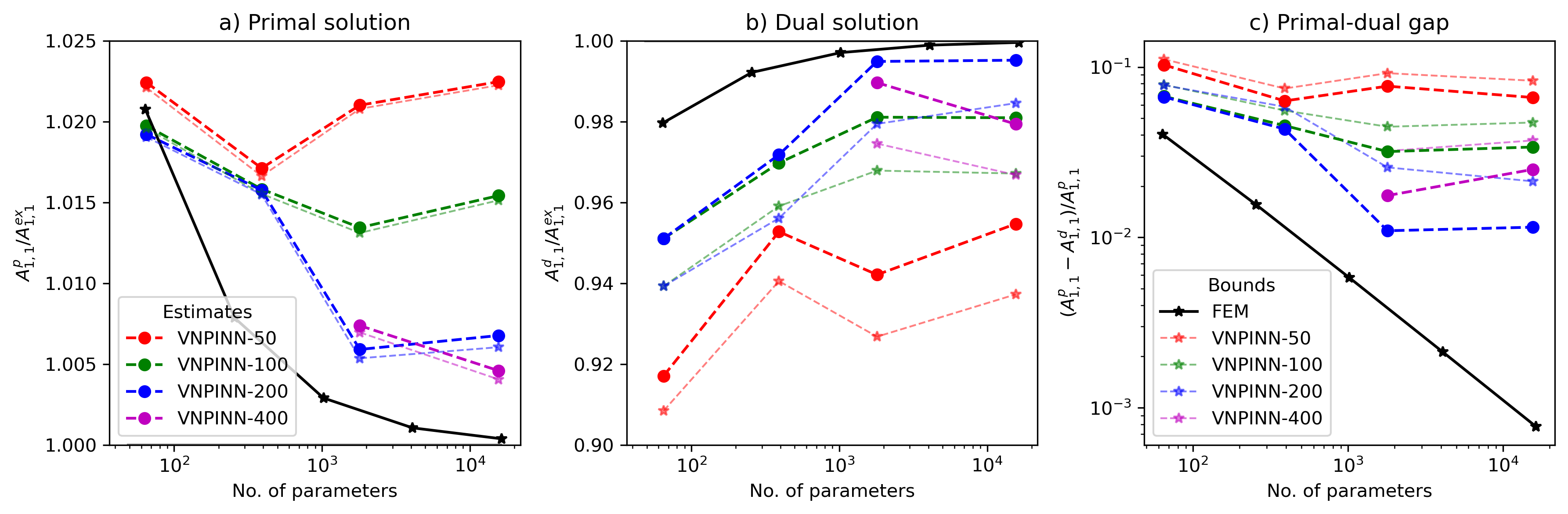}
    \caption{Precision of the primal and dual estimates of the VNPINN depending on the number of network parameters and the number of neural network test functions.}
    \label{fig:vnpinn-estimates}
\end{figure}

Four differently sized PINN architectures using $N_t=\{50, 100, 200\}$ were trained. The networks with $1\,801$ and $15\,601$ parameters were additionally trained with 400 test functions. The primal and dual estimates, as well as the guaranteed bounds, are shown in Fig.~\ref{fig:vnpinn-estimates}.

\begin{figure}[h]
    \centering
    \includegraphics[width=1\linewidth]{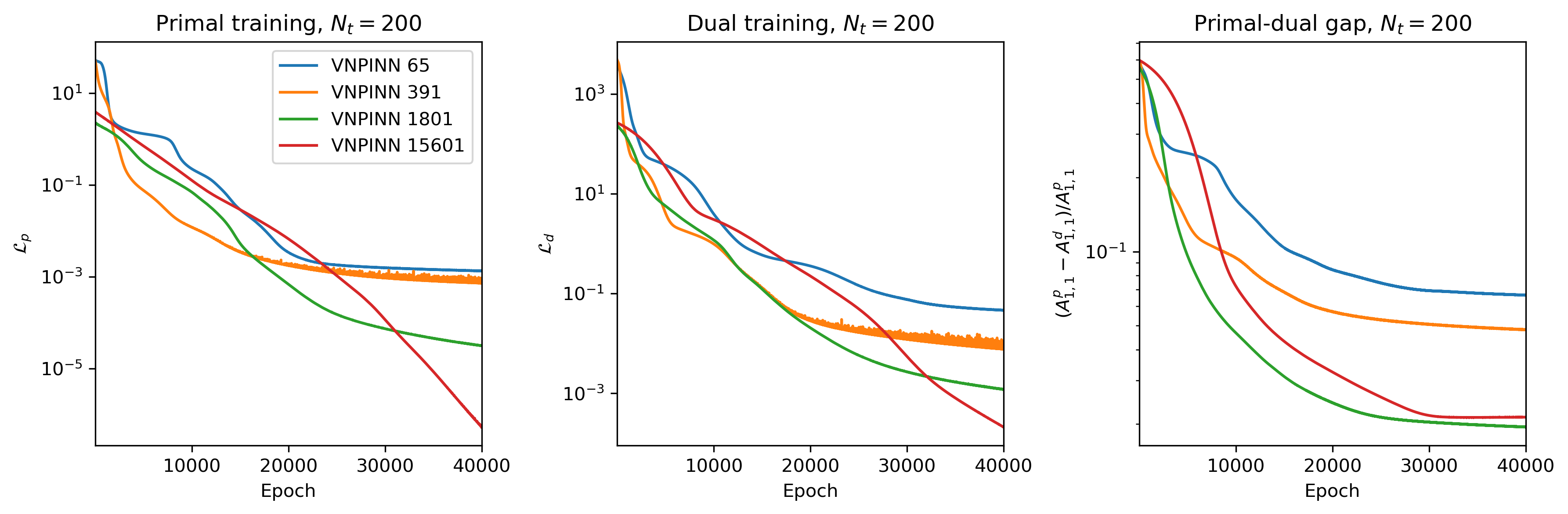}
    \caption{Training loss of the primal and dual solutions, and the gap between the estimates, VPINN with 200 neural network test functions (VNPINN).}
    \label{fig:training200-vnpinn}
\end{figure}

\begin{figure}[h]
    \centering
    \includegraphics[width=1\linewidth]{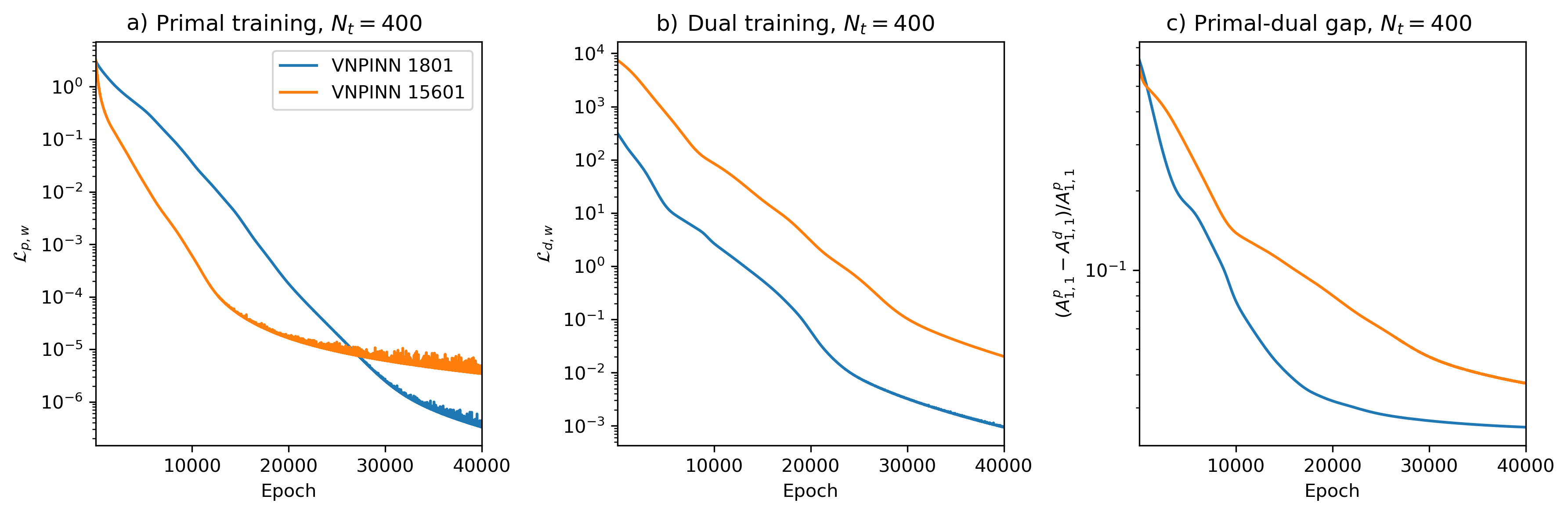}
    \caption{Training loss of the primal and dual solutions, and the gap between the estimates, VPINN with 200 neural network test functions (VNPINN).}
    \label{fig:training400-vnpinn}
\end{figure}

\begin{figure}[h]
    \centering
    \includegraphics[width=1\linewidth]{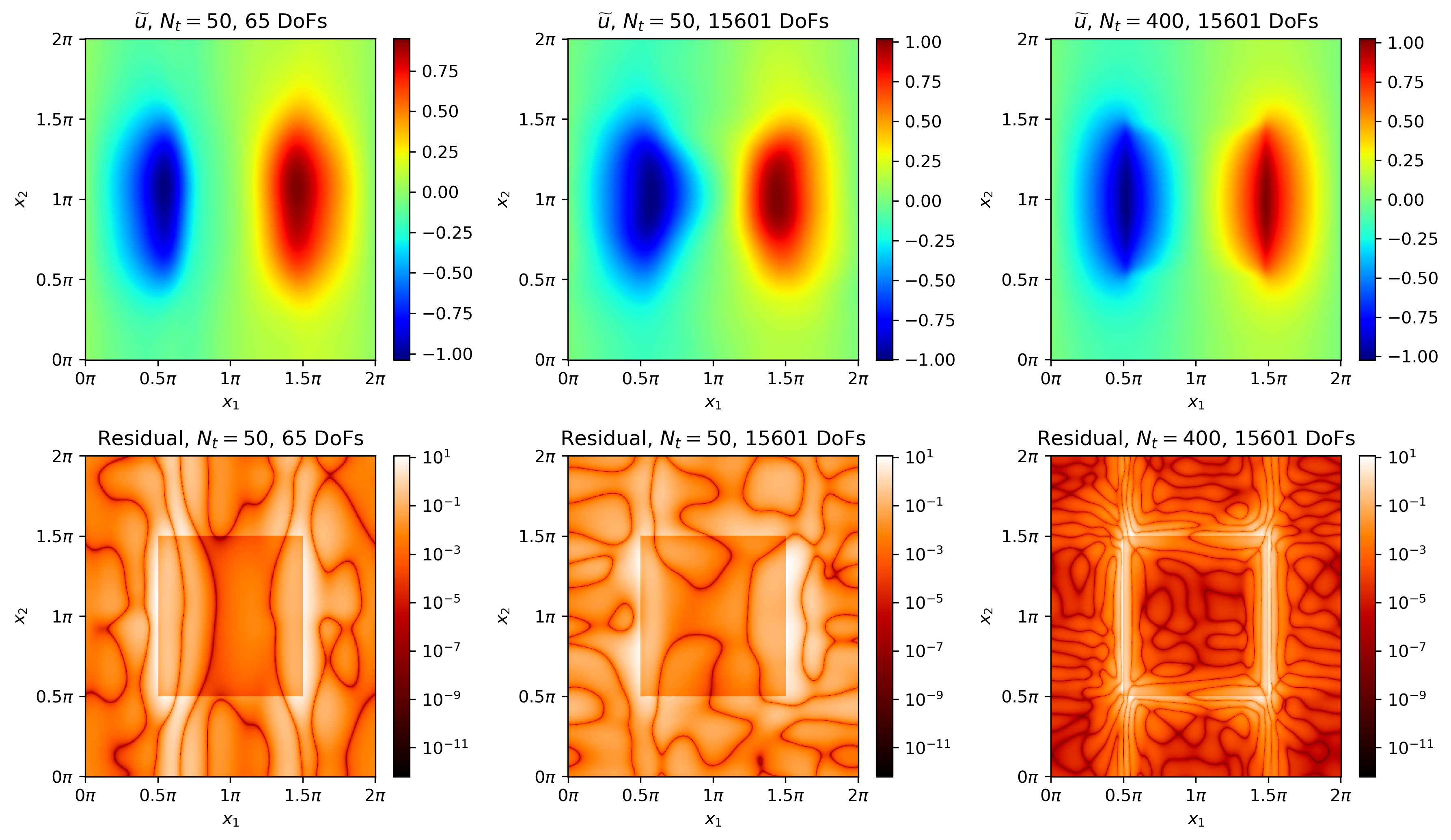}
    \caption{Primal solutions (top) and the residuals (bottom) of select VNPINNs depending on the number of the neural network test functions $N_t$.}
    \label{fig:vnpinn-primal}
\end{figure}

\begin{figure}[h]
    \centering
    \includegraphics[width=1\linewidth]{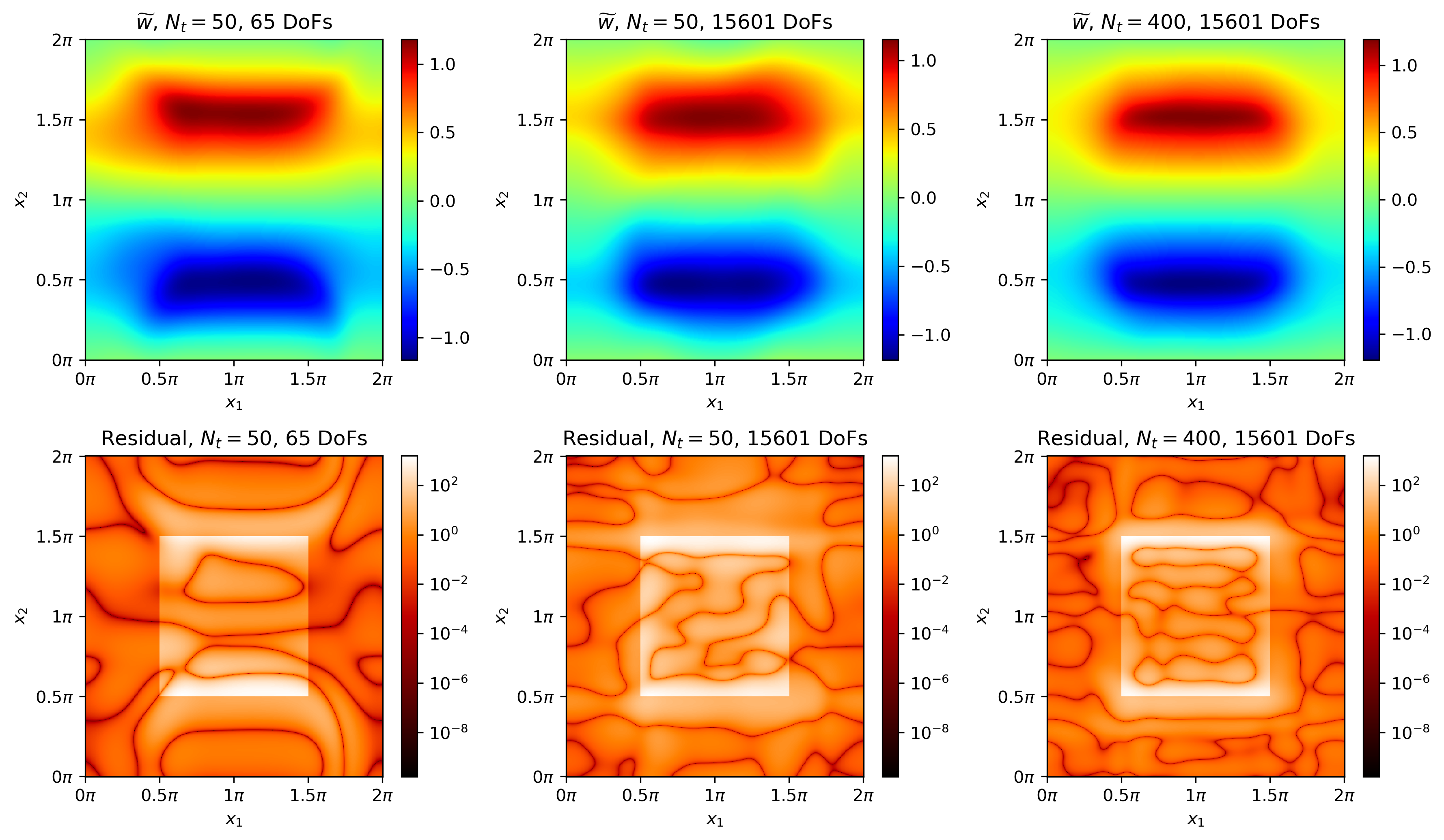}
    \caption{Dual solutions (top) and the residuals (bottom) of select VNPINNs depending on the number of the neural network test functions $N_t$.}
    \label{fig:vnpinn-dual}
\end{figure}

Here, we observe that both primal and dual solutions improve with the increasing number of parameters and test functions. Increasing the number of test functions for the smaller networks (65 and 391 parameters) has diminishing returns, as these networks perform similarly with 100 and 200 test functions. On the other hand, the precision of larger networks does not improve if the number of test functions does not increase accordingly. For example, the VNPINNs with $15\,601$ parameters trained with 50 test functions yield worse effective conductivity estimates and have higher strong-form residuals than the VNPINNs with 65 parameters trained with the same number of test functions. Since the achieved training loss of the large network is generally lower, the distortions in the solutions in Figs.~\ref{fig:vnpinn-primal} and \ref{fig:vnpinn-dual} for the solution with $15\,601$ parameters and $N_t=50$ suggest the overfitting to the test functions. \revision{While this negative effect is less pronounced when the number of test functions is large (i.e., $N_{t}\in{200, 400}$), the comparison between the training loss and the primal-dual gap in Figs.~\ref{fig:training200-vnpinn}~and~\ref{fig:training400-vnpinn}) shows that while the convergence in training loss has not been reached in the allocated $40\,000$ epochs, the primal-dual gap has reached convergence around $30\,000$-th epoch, which, again, suggests the overfitting to the test functions.}

From the presented results, we conclude that the large networks ($1\,801$ and $15\,601$ parameters) gain around 1\% precision of the primal estimate for each doubling of the test basis, and up to 4\% in the dual estimate. The point of diminishing returns is reached around 400 test functions, where the only improvement is in the primal solution by the largest network. This is explained by the overall slowdown in training for such a large number of test functions (Fig.~\ref{fig:training400-vnpinn}), compared to the smaller test basis (Fig.~\ref{fig:training200-vnpinn}). 

\FloatBarrier

\subsection{Multiple Inclusions}
\label{sec:multiple_inc}
\begin{figure}[h]
    \centering
    \includegraphics[width=0.8\linewidth]{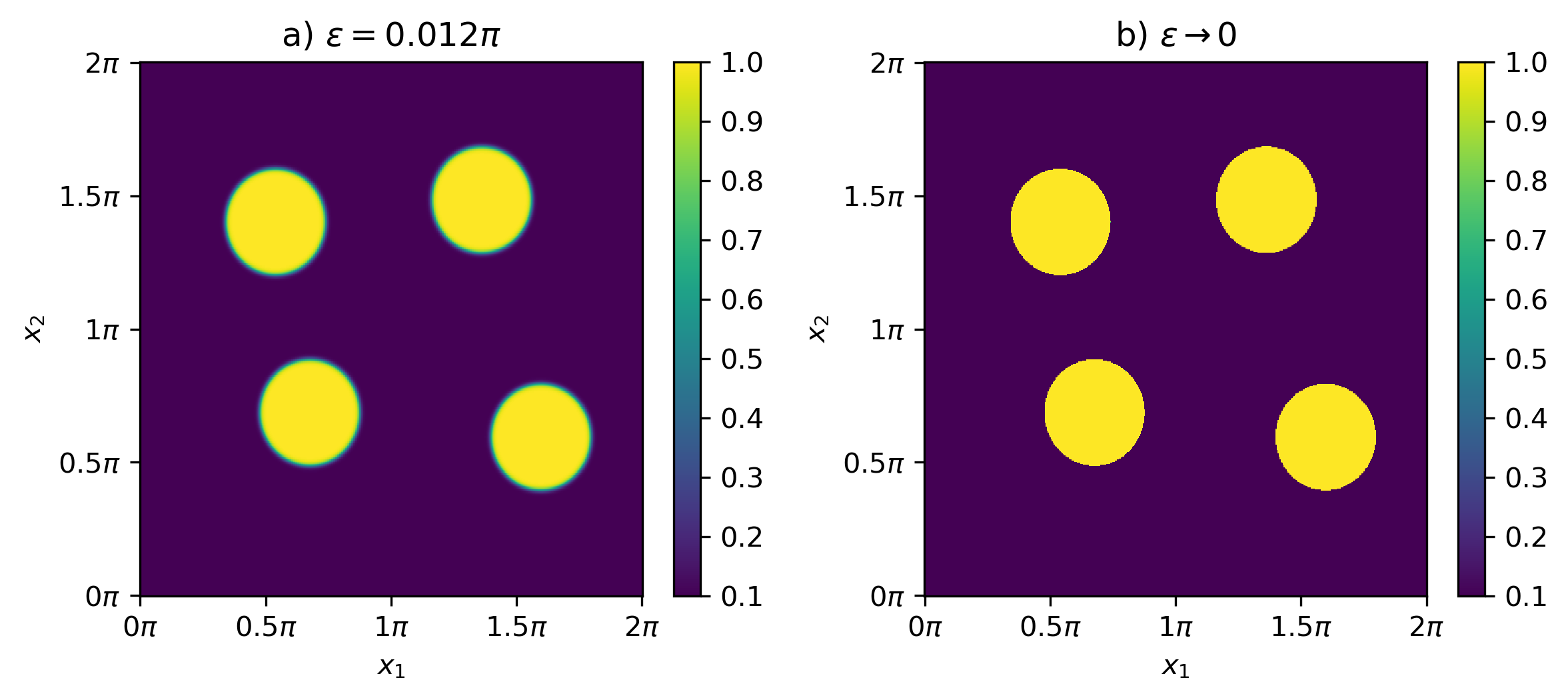}
    \caption{\revision{Smooth approximation (a) $\bm{A}_{s,\varepsilon}(\bm{x})$ of the material with multiple inclusions (b)}}
    \label{fig:material_multi} 
\end{figure}

\revision{The performance of strong-form PINNs in the primal formulation was thoroughly tested by \cite{JIANG2023115972} on multi-inclusion composites with varying volume fraction, where it was shown to achieve the same precision in the homogenized parameters as FEM. However, the aforementioned study did not consider the error introduced by the material smoothing. Here, we assume a composite with four randomly distributed equi-sized inclusions in a $2\pi \times 2\pi$ periodic cell, where $\gamma_{\mathrm{mat}}=0.1$ and $\gamma_{\mathrm{inc}}=1$ are thermal conductivities of the matrix and the inclusions, respectively, see Fig.~\ref{fig:material_multi}. For the PINN solution, we assume the smooth approximation of the conductivity function as
\begin{equation}
    \bm{A}_{s,\varepsilon}(\bm{x}) = \left(0.55 - 0.45 \prod_{i=1}^{4} \tanh \left[\frac{1}{\varepsilon}(||\bm{x}-\bm{x}^c_i||^2_2 -r^2)\right]\right)\bm{I},
\end{equation}
where $\bm{x}^c_i$ are the centers of the inclusions, $r$ is the radius of the inclusions, and $\varepsilon > 0$ controls the sharpness of the transition. Since the previous example with a single inclusion showed that there is an optimal $\varepsilon$ that minimizes the material approximation error while still allowing the PINN to converge to a high-quality solution, we only test the PINNs on one such $\varepsilon$, which we found to be roughly equal to 0.06 of the inclusion radius $0.2\pi$, or $0.012 \pi$. 
}

\begin{figure}[htbp]
    \centering
    \includegraphics[width=0.8\linewidth]{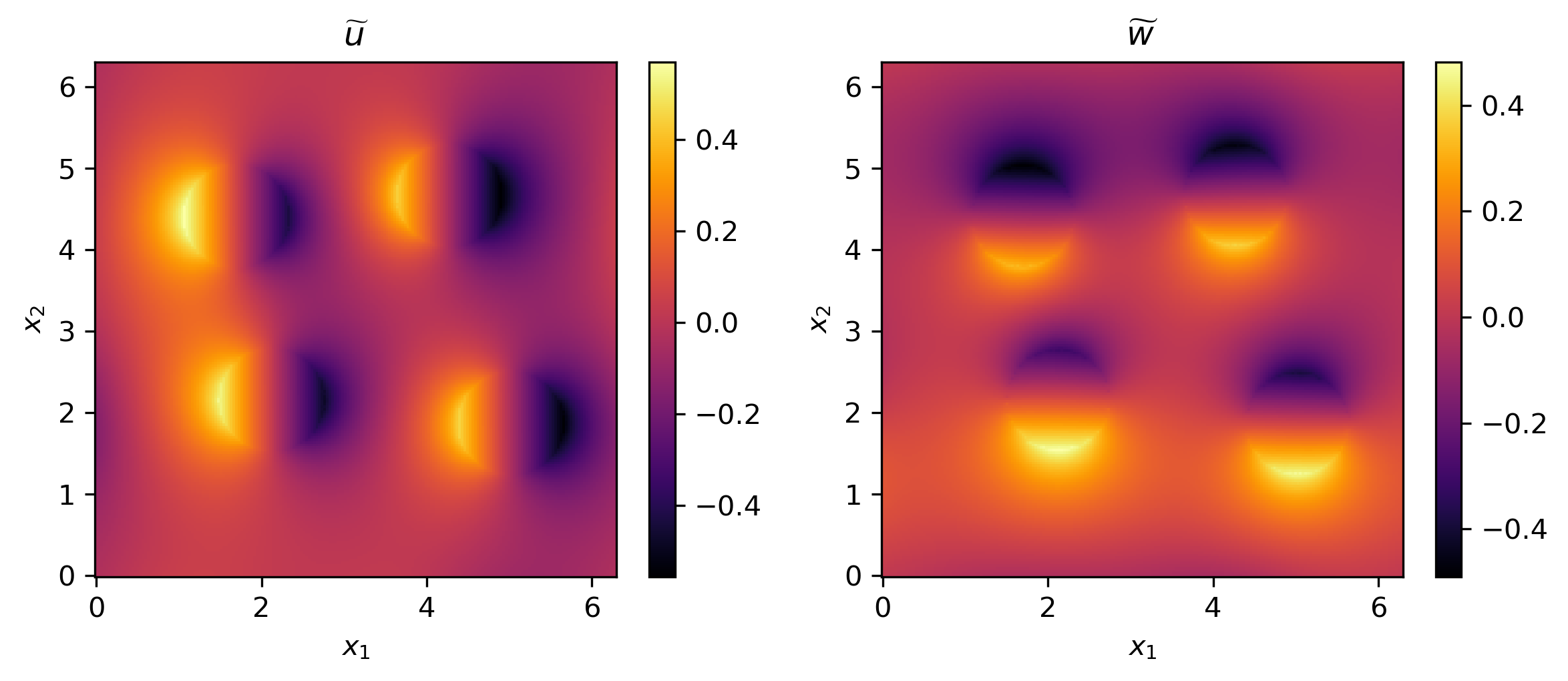}
    \caption{\revision{Benchmark primal and dual solutions for $\bm{\xi}=[1,0]^{\rm{T}}$, obtained with linear triangular finite elements, using discretization into $200 \times 200$ unique DoFs.}}
    \label{fig:fem_multi} 
\end{figure}

\revision{As there is no known analytical solution for this problem, we compare the neural network solutions only to the FEM benchmark computed on a $200 \times 200$ regular mesh, 
which provides the upper and lower bounds within 0.35\% of each other, see Fig.~\ref{fig:fem_multi}. We then test two larger configurations of the neural networks (Tab.~\ref{tab:NN_config_2}) in the primal and dual formulation as PINNs with the smooth material approximations and VNPINNs with 400 test functions. The collocation points are selected on the $200 \times 200$ regular grid, which is then used for trapezoidal integration. All networks were trained for $100\,000$ epochs using the Adam optimizer and the initial learning rate of $10^{-5}$.
}

\begin{table}[h]
    \centering
    \begin{tabular}{|c|c|c|c|}
        \hline
        \footnotesize{\textnumero~neurons, periodic layer} & \footnotesize{\textnumero~neurons, per residual layer} & \footnotesize{\textnumero~residual layers} & \footnotesize{\textnumero~parameters}\\
        \hline
        50 & 50 & 5 & $15\,601$\\
        \hline
        70 & 70 & 6 & $35\,211$\\
        \hline
    \end{tabular}
    \caption{\revision{Configurations of the considered neural networks for both primal versus dual and strong versus weak formulations, for the multiple inclusions problem.}}
    \label{tab:NN_config_2}
\end{table}

\begin{figure}[h]
    \centering
    \includegraphics[width=1\linewidth]{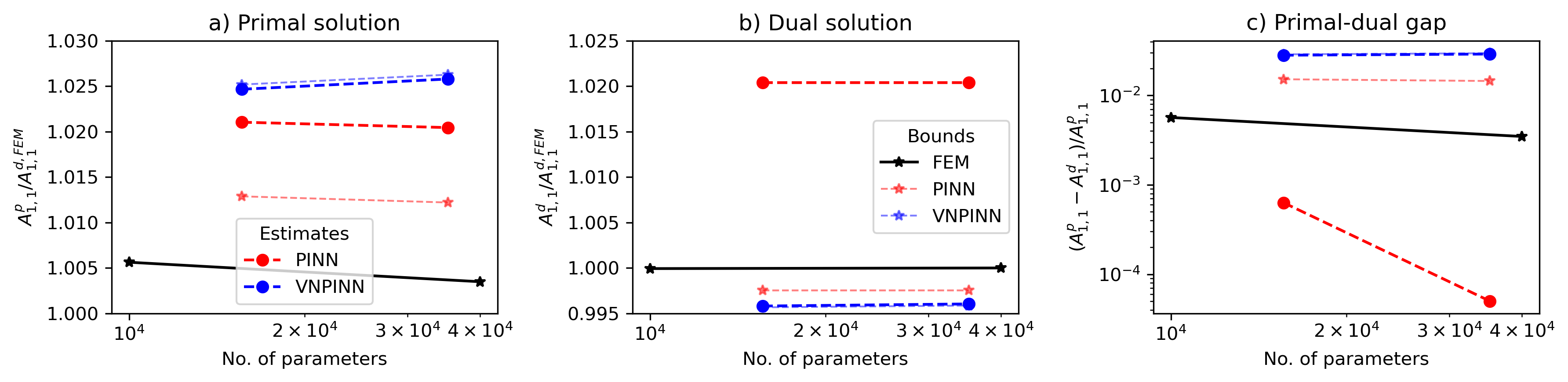}
    \caption{\revision{Precision of the primal and dual estimates of the PINN and VNPINN depending on the number of network parameters.}}
    \label{fig:estimates_multi}
\end{figure}

\begin{figure}[h]
    \centering
    \includegraphics[width=1\linewidth]{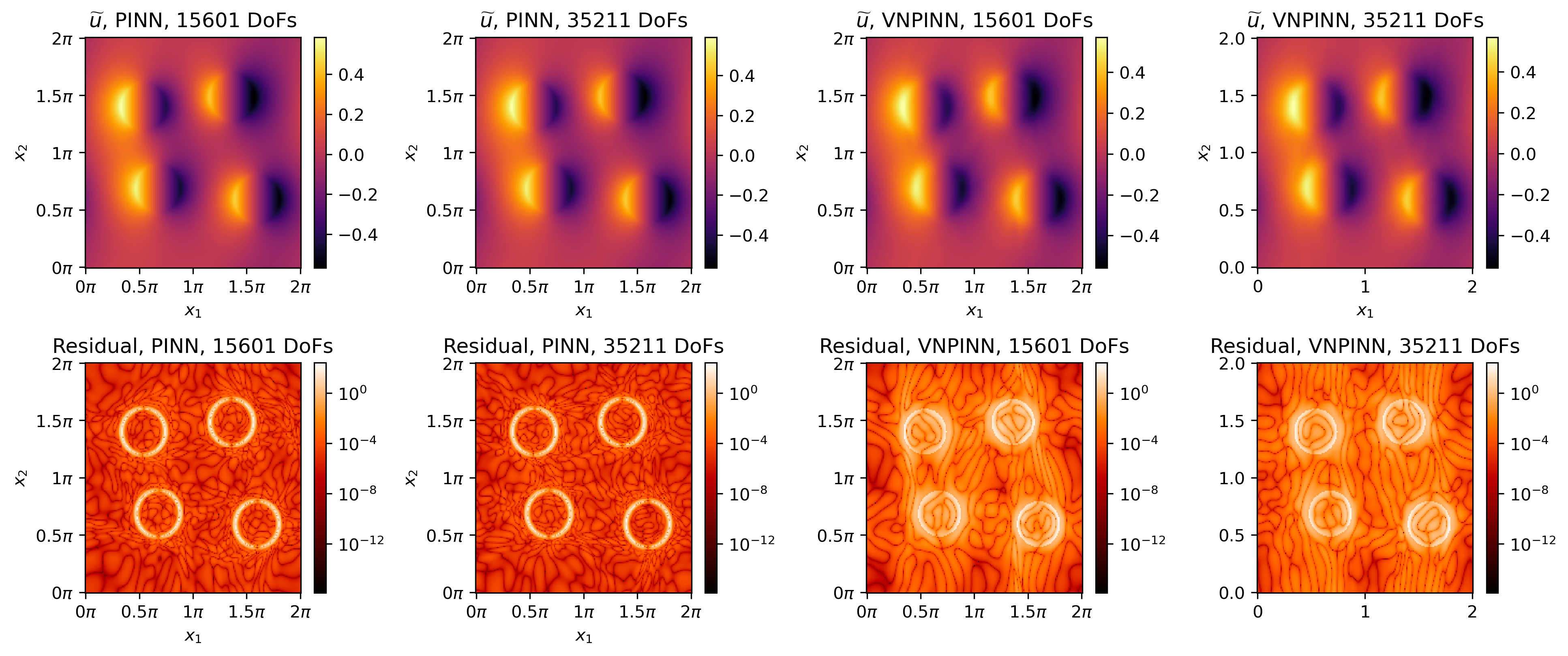}
    \caption{\revision{Primal solutions (top) and the residuals (bottom) of PINNs and VPINNs for the multiple inclusion problem.}}
    \label{fig:primal_multi}
\end{figure}

\begin{figure}[h]
    \centering
    \includegraphics[width=1\linewidth]{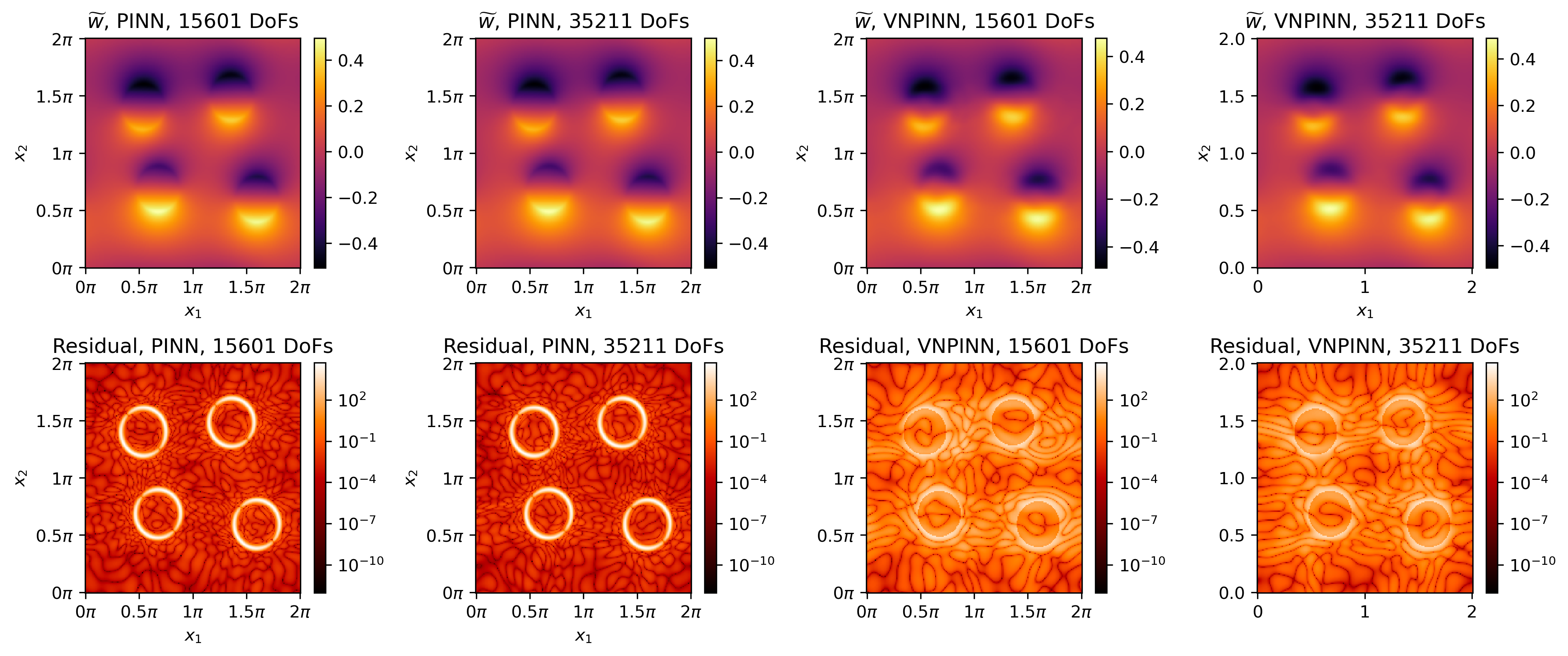}
    \caption{\revision{Dual solutions (top) and the residuals (bottom) of PINNs and VPINNs for the multiple inclusion problem.}}
    \label{fig:dual_multi}
\end{figure}

\revision{
The estimates and bounds of the homogenized parameter $A_{1,1}^*$ from the primal and dual solutions for PINNs and VPINNs are shown in Fig.~\ref{fig:estimates_multi}, normalized by the dual FEM bound. Both $15\,601$ and $35\,211$-parameter PINN converged to virtually the same parameter estimate in the primal and dual formulation. However, when the material approximation error is considered, PINNs provide an estimate around 2\% higher than the FEM benchmark, while the calculated bounds on the exact material distribution are up to 1.2\% away from the FEM benchmark. Even with the material approximation error, PINNs have achieved a closer result to the benchmark than the VNPINNs that have been trained on the exact material distribution, whose primal bounds and estimates come to around 2.5\% higher than the benchmark, and dual bounds and estimates are 0.5\% lower. The PINN and VNPINN solutions and their residuals can be compared in Figs.~\ref{fig:primal_multi} and ~\ref{fig:dual_multi}. Interestingly, unlike in the previous example, the dual formulation provided a more accurate estimate. One possible explanation could lie in the opposite contrast ratios of selected conductivities in the matrix and inclusion in the two examples. The dependence on the relative conductivity might be corroborated by the observation that the strong-form residuals of the dual VPINNs are significantly higher for the less conductive (i.e., more resistant) square inclusion in comparison to the matrix in Fig.~\ref{fig:vnpinn-dual}, while more conductive circular inclusions have higher residuals in the primal formulation in Fig.~\ref{fig:dual_multi}. Strong-form PINNs typically only have significantly higher residuals at the interface of the materials. The behavior of VPINNs for different material contrasts should be explored in a future study.
}

\begin{figure}[h]
    \centering
    \includegraphics[width=1\linewidth]{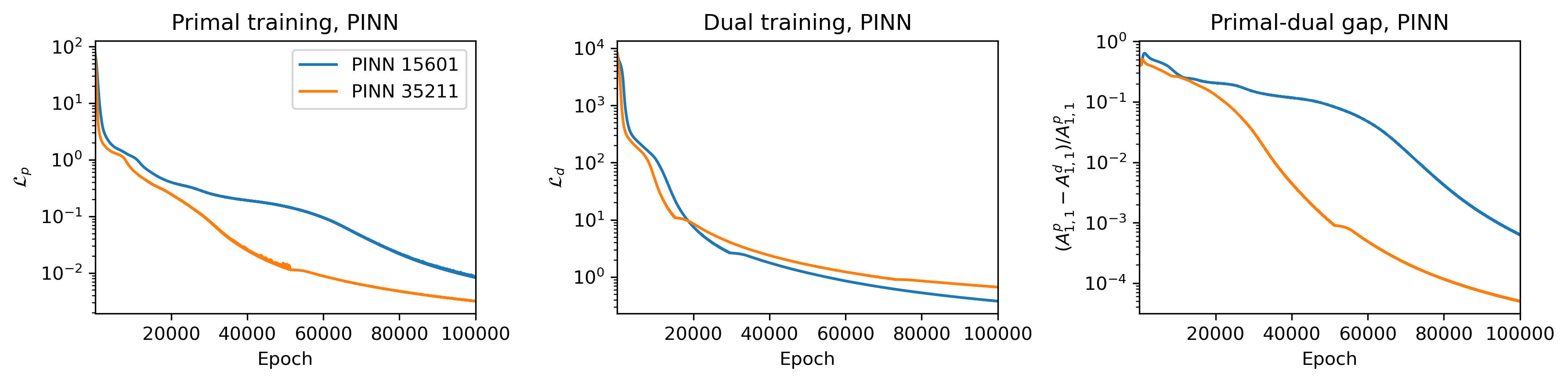}
    \caption{\revision{Training loss of the primal and dual solutions, and the gap between the estimates, PINN, multiple inclusion problem.}}
    \label{fig:pinn_training_multi}
\end{figure}

\begin{figure}[h]
    \centering
    \includegraphics[width=1\linewidth]{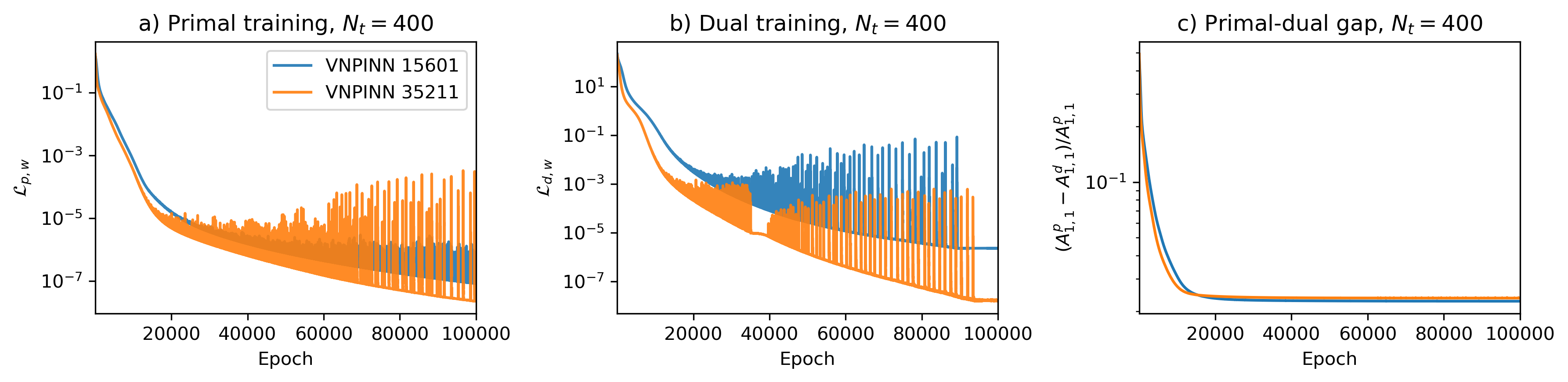}
    \caption{\revision{Training loss of the primal and dual solutions, and the gap between the estimates, VNPINN, multiple inclusion problem.}}
    \label{fig:vpinn_training_multi}
\end{figure}

\revision{The training of PINNs and VPINNs is shown in Fig.~\ref{fig:pinn_training_multi} and Fig.~\ref{fig:vpinn_training_multi}. The VPINN estimates stopped improving after about $20\,000$ epochs, which suggests the overfitting to the test functions, while PINN estimates achieved a similar precision slightly slower (around $30\,000$ epochs) but continued to improve significantly. However, considering that material smoothing puts a hard limit on the guaranteed bounds, the utility of such a prolonged training of PINNs should be assessed. }

\FloatBarrier

\section{Discussion}\label{sec:discussion}

The relative error of the final primal and dual estimates for all networks \revision{of the single inclusion experiment (Section \ref{sec:single_inclusion})} is summarised in Tab.~\ref{tab:error}. Tab.~\ref{tab:bounds} then lists the computed guaranteed error bounds from the piecewise linear approximations of the solutions. 

\begin{table}[h]
    \centering
    \begin{tabular}{|c|rrrr|rrrr|}
        \hline
        Formulation & \multicolumn{4}{c|}{Primal} & \multicolumn{4}{c|}{Dual} \\
        \hline
        N. of parameters & 65 & 391 & $1\,801$ & $15\,601$ & 65 & 391 & $1\,801$ & $15\,601$ \\
        \hline
        \hline
        PINN, $\varepsilon=1/10$ & \bf{1.944} & 1.442 & 1.441 & 1.441 & \bf{0.167} & 1.424 & 1.440 & 1.441\\
         
        PINN, $\varepsilon=1/20$ & 10.772 & 0.695 & 0.694 & 0.694 & -12.312 & \bf{0.544} &   0.671 & 0.693 \\
         
        PINN, $\varepsilon=1/30$ & 6.459 & \bf{0.461} & \bf{0.456} & \bf{0.457} & -26.174 & -7.417 & \bf{-0.184} & \bf{0.448}\\

        PINN, $\varepsilon=1/40$ & 15.518 & 20.103 & 19.686 & 11.507 & -24.564 & -16.308 & -44.057 & -32.990\\
        \hline
        \hline
        VSPINN, $N_t=70$ & 2.136 & 2.713 & 2.333 & 3.054 & \bf{-3.819} & -7.229 & -5.820 & -6.388\\
         
        VSPINN, $N_t=126$ & 1.945 & 1.779 & 4.734 & 7.054 & -5.023 & -6.010 & -7.937 & -6.091\\
        \hline
        VNPINN, $N_t=50$ & 2.243 & 1.712 & 2.102 & 2.247 & -8.288 & -4.722 & -5.786 & -4.527 \\
         
        VNPINN, $N_t=100$ & 1.975 & 1.583 & 1.345 & 1.543 & -4.889 & -3.021 & -1.889 & -1.902\\
         
        VNPINN, $N_t=200$ & \bf{1.922} & \bf{1.577} & \bf{0.591} & 0.677 & -4.889 & \bf{-2.817} & \bf{-0.509} & \bf{-0.478}\\
         
        VNPINN, $N_t=400$ & --- & --- & 0.739 & \bf{0.460} & --- & --- & -1.035 & -2.059\\
        \hline
    \end{tabular}
    \caption{Relative error (\%) of the primal and dual {\bf estimates} of the homogenized parameter for differently trained networks,  best result in each group is in \bf{bold}.}
    \label{tab:error}
\end{table}

\begin{table}[h]
    \centering
    \begin{tabular}{|c|rrrr|rrrr|}
        \hline
        Formulation & \multicolumn{4}{c|}{Primal} & \multicolumn{4}{c|}{Dual} \\
        \hline
        N. of parameters & 65 & 391 & $1\,801$ & $15\,601$ & 65 & 391 & $1\,801$ & $15\,601$ \\
        \hline
        \hline
        PINN, $\varepsilon=1/10$ & \bf{1.190} & 0.717 & 0.718 & 0.718 & \bf{-4.922} & -4.465 & -4.490 & -4.493 \\
         
        PINN, $\varepsilon=1/20$ & 10.460 &  0.359 &  0.359 &  0.359 & -12.883 &  \bf{-2.068} &  -2.065 &  -2.064 \\
         
        PINN, $\varepsilon=1/30$ & 6.204 & \bf{0.246} & \bf{0.240} & \bf{0.244} & -26.218 &  -7.990 &  \bf{-1.608} &  \bf{-1.168}\\

        PINN, $\varepsilon=1/40$ & 14.729 & 18.727 & 19.501 & 11.195 & -24.578 & -18.093 & -45.371 & -33.549\\
        \hline
        \hline
        VSPINN, $N_t=70$ & 2.099 & 2.692 & 2.298 & 3.033 & \bf{-5.013} & -8.569 & -7.003 & -7.734\\
         
        VSPINN, $N_t=126$ & 1.912 & 1.749 & 4.707 & 7.035 & -6.288 & -7.183 & -8.883 & -7.551\\
        \hline
        VNPINN, $N_t=50$ & 2.211 & 1.663 & 2.080 & 2.226 & -9.151 & -5.944 & -7.319 & -6.272 \\
         
        VNPINN, $N_t=100$ & 1.959 & 1.554 & 1.312 & 1.514 & -6.069 & \bf{-4.089} & -3.210 & -3.282\\
         
        VNPINN, $N_t=200$ & \bf{1.905} & \bf{1.550} & \bf{0.537} & 0.605  & -6.069 & -4.388 & \bf{-2.053} & \bf{-1.542}\\
         
        VNPINN, $N_t=400$ & --- & --- & 0.697 & \bf{0.406} & --- & --- & -2.542 & -3.317\\
        \hline
    \end{tabular}
    \caption{Relative error (\%) of guaranteed primal and dual {\bf bounds} of the homogenized parameter for differently trained networks, best result in each group is in \bf{bold}.}
    \label{tab:bounds}
\end{table}

As shown in Tab.~\ref{tab:bounds}, the strong-form PINNs demonstrate superior convergence for smoothly approximated materials with $\varepsilon \in \{1/10, 1/20, 1/30\}$, achieving relative errors below 0.5\% for the largest networks in the primal formulation. However, this performance degrades catastrophically when $\varepsilon = 1/40$, with relative errors exceeding 10\% even for the $15\,601$-parameter network, indicating a critical failure mode when the material transition becomes too sharp.

Among the variational approaches, VNPINNs with neural network test functions outperform VSPINNs for most configurations examined. The VSPINNs exhibit non-monotonic behavior with respect to network size, as evidenced by the degraded performance of the 15,601-parameter network with 126 test functions (7.054\% error) compared to smaller networks. This suggests overfitting to integration errors when high-frequency spectral basis functions are employed with insufficient quadrature precision. In contrast, VNPINNs show more consistent improvement with increasing \revision{the number of} test functions, achieving their best performance with 400 test functions for the largest networks, reaching 0.460\% relative error in the primal formulation.

The guaranteed bounds presented in Tab.~\ref{tab:bounds} are close to the estimates, with the projection onto piecewise linear finite elements introducing minimal additional error (typically less than 0.5\%). \revision{With precise integration,} the bounds calculated from the projected solution maintain the theoretical ordering property of Eq.~\eqref{eq:order} for all solutions, providing a reliable indicator of solution quality. \revision{Unfortunately, this requires a careful meshing of the geometry, which is not always feasible and diminishes one of the primary benefits of PINNs. However, in practice, a reasonably precise numerical integration typically produces reliable results as well, as demonstrated by the primary estimates that can be computed fast during the training of the network.}

\revision{For the multiple inclusions example, we similarly conclude that strong-form PINNs can perform extremely well for smooth geometries, but cannot match even simple piece-wise linear FEM for discontinuous materials. The quality of the test basis remains the main limiting factor for VPINNs, and the number of required test functions depends on the complexity of the geometry as well.}

\section{Conclusion}\label{sec:conclusion}

This work demonstrates the integration of dual formulation principles from classical homogenization theory into the PINN framework for computing effective properties of thermoconductive composites. The dual approach provides not only theoretically guaranteed bounds on homogenized parameters but also serves as a practical diagnostic tool for identifying convergence failures that might otherwise go undetected in standard PINN implementations.

We examined the effectiveness of dual formulations within two primary strategies for addressing PINN limitations with discontinuous materials: (i) smooth approximation of material coefficients in strong-form PINNs, and (ii) variational formulations with both spectral and neural network test function bases. Our findings reveal fundamental trade-offs between these approaches. Strong-form PINNs achieve superior accuracy for sufficiently smooth material approximations but require careful calibration of the smoothing parameter to avoid catastrophic\revision{(and unidentifiable without the duality gap calculation )} failure. VPINNs, though computationally more demanding and sensitive to test function selection, offer greater robustness by directly accommodating discontinuous material properties.

The primal-dual framework enhances the reliability of PINN-based homogenization by providing complementary solutions that bound the true effective properties. The gap between primal and dual estimates serves as an intrinsic quality metric, enabling practitioners to assess solution reliability without reference solutions.

Looking forward, extending this dual formulation approach to three-dimensional problems and linear elasticity presents both challenges and opportunities. The divergence-free constraint in 3D cannot be represented through a scalar stream function as in 2D, necessitating modified network architectures and additional training loss terms. However, classical numerical methods face similar increases in complexity in their dual formulations, requiring significantly more degrees of freedom than their primal counterparts. This parallel suggests that PINN-based approaches may retain competitive advantages even as problem complexity increases, particularly given their mesh-free nature and potential for rapid evaluation once trained.

Implementation and trained neural networks are available on \href{https://github.com/LiyaGaynutdinova/Homogenization-VPINN}{GitHub}.

\section*{CRediT Author Statement}
LG: Conceptualization, Methodology, Software, Investigation, Writing -- Original Draft, Visualization, Funding Acquisition; 
OR: Methodology, Writing -- Review \& Editing, Supervision;
MD: Methodology, Writing -- Review \& Editing, Supervision;
IP: Methodology, Writing -- Review \& Editing, Supervision, Funding Acquisition.

\section*{Acknowledgments}
LG and IP acknowledge funding by the European Union under the project ROBOPROX (reg. no. CZ.02.01.01/00/22\_008/0004590). MD's work was supported by the Czech Science Foundation, project No.~24-11845S. LG also acknowledges the Student Grant Competition of the Czech Technical University in Prague (project no.~SGS25/002/ OHK1/1T/11). 

\FloatBarrier

\bibliographystyle{elsarticle-harv} 
\bibliography{cas-refs}

\end{document}